\documentclass[twoside,11pt]{article}
\usepackage[nottoc]{tocbibind}
\usepackage[usehyper]{jmlr2e-for-arXiv}
\hypersetup{colorlinks=true, allcolors=blue, bookmarks=true}
\usepackage{amsmath}

\usepackage{enumerate}
\usepackage{array}
\usepackage{url} 
\usepackage{rotating}
\usepackage{placeins}
\usepackage{multirow}
\usepackage{bbm}
\usepackage{doi}
\usepackage{algorithm}
\usepackage{algorithmicx}
\usepackage{mathtools}
\usepackage{graphicx}
\usepackage{float}
\usepackage{color, soul} 
\usepackage[none]{hyphenat}

\makeatletter
\@fpsep\textheight
\makeatother

\usepackage{mathtools} 
\usepackage{tikz, tikz-qtree}
\usetikzlibrary{positioning}
\usetikzlibrary{arrows}
\usepackage[framemethod=TikZ]{mdframed}

\newtheorem{assumption}{Assumption}
\newtheorem{rmrk}{Remark}
\def\trans{^{\rm T}} 
\newcommand{\x}{{\bf x}}

\newcommand{\X}{{\bf X}}
\newcommand{\w}{{\bf w}}

\newcommand{\cF}{\mathcal{F}}
\newcommand{\cG}{\mathcal{G}}

\newcommand{\Cov}{\mathrm{Cov}}

\newcommand{\hide}[1]{}

\usepackage{scalerel,stackengine}
\stackMath
\newcommand\reallywidehat[1]{%
\savestack{\tmpbox}{\stretchto{%
  \scaleto{%
    \scalerel*[\widthof{\ensuremath{#1}}]{\kern-.6pt\bigwedge\kern-.6pt}%
    {\rule[-\textheight/2]{1ex}{\textheight}}
  }{\textheight}%
}{0.5ex}}%
\stackon[1pt]{#1}{\tmpbox}%
}

\definecolor{matlab_blue}{RGB}{0,114,189}
\definecolor{matlab_red}{RGB}{217, 83, 25}

\DeclareMathOperator*{\argmin}{\arg\!\min}

\DeclareMathOperator*{\E}{\mathbb{E}}
\DeclarePairedDelimiterX{\inp}[2]{\langle}{\rangle}{#1, #2}

\jmlrheading{*}{2020}{*-**}{*/**}{*/**}{Tenzer}{Yaniv Tenzer, Amit Moscovich, Mary Frances Dorn, Boaz Nadler and Clifford Spiegelman}

\ShortHeadings{Beyond Trees: Classification with Sparse Pairwise Dependencies}{Tenzer, Moscovich, Dorn, Nadler and Spiegelman}
\firstpageno{1}

\begin{document}

\title{Beyond Trees:  Classification with Sparse Pairwise~Dependencies}

\author{\name Yaniv Tenzer \email yaniv.tenzer@weizmann.ac.il\\
    \addr Department of Computer Science and Applied Mathematics\\
    Weizmann Institute of Science \\
    Rehovot, 76100, Israel
    \AND    
    \name Amit Moscovich
\email amit@moscovich.org \\
    \addr Program in Applied and Computational Mathematics\\
    Princeton University\\
    Princeton, NJ 08544, USA
    \AND
    \name Mary Frances Dorn \email mfdorn@lanl.gov \\
    \addr Los Alamos National Laboratory\\
    P.O. Box 1663, MS F600\\
    Los Alamos, NM 87545, USA
    \AND       
    \name Boaz Nadler \email boaz.nadler@weizmann.ac.il\\
    \addr Department of Computer Science and Applied Mathematics\\
    Weizmann Institute of Science \\
    Rehovot, 76100, Israel
    \AND
    \name Clifford Spiegelman \email cliff@stat.tamu.edu \\
    \addr Department of Statistics\\
    Texas A\&M University\\
    College Station, TX 77843, USA
}

\editor{TBD}

\maketitle

\begin{abstract}
Several classification methods assume that the underlying distributions follow tree-structured graphical models. 
Indeed, trees capture statistical dependencies between pairs of variables, which may be crucial to attain low classification errors.
The resulting classifier is linear in the log-transformed univariate and bivariate densities that correspond to the tree edges. 
In practice, however, observed data may not be well approximated by trees. 
Yet, motivated by the importance of pairwise dependencies for accurate classification, 
here we propose to 
approximate the optimal decision boundary by a sparse linear combination of the univariate and bivariate log-transformed densities. 
Our proposed approach is semi-parametric in nature: we non-parametrically estimate the univariate and bivariate densities, remove pairs of variables 
that are nearly independent using the Hilbert-Schmidt independence criteria, and finally construct  a linear SVM on the retained log-transformed densities. 
We demonstrate using both synthetic and real data that our resulting classifier, denoted SLB (Sparse Log-Bivariate density), 
is competitive with popular classification methods. 
\end{abstract}


%
%
\section{Introduction}
\label{sec:intro}


Consider a binary classification problem
where the vector of explanatory variables \({\X} = (X_1, \ldots, X_d)\) belongs to some \(d\)-dimensional space 
$\Omega\subseteq \mathbb{R}^d$, and the response variable $Y$ takes values in the set $\mathcal Y=\{+1,-1\}$. 
Given a training set of labeled samples $\{{\bf x}_{\ell},y_{\ell}\}_{\ell=1}^n$, the goal is to construct a classifier \(f:\Omega\to \mathcal Y\), with a small misclassification error rate on future unlabeled instances of $\bf X$. 

It is well known that the optimal classifier is a threshold of the likelihood ratio statistic,
\begin{equation} \label{eq:LR}
\text{LR}({\bf x}) = \frac{p(\x|Y=+1)}{p(\x|Y=-1)}.
\end{equation}
A key challenge in applying this result, is the estimation of the class-conditional densities $p(\x|Y=y)$. Unfortunately, accurate non-parametric density estimation requires a number of labeled samples that is exponential in the dimension $d$.
This is known as the \emph{curse of dimensionality} (see  chapter 2 of \cite{Tsybakov2009}). Hence, in high dimensional settings with a limited number of labeled samples, classification methods based on non-parametric multivariate density estimates may perform poorly, and are not widely used in practice.

Instead of a non-parametric estimate of $p(\x|y)$, a popular alternative is to model this class conditional multivariate density with  Bayesian Networks (BNs) \citep{Lauritzen1996}. BNs provide a powerful probabilistic representation of multivariate distributions, which in turn often yields accurate classifiers. 
However, learning an unrestricted BN may require a large number of labeled samples, and moreover may be computationally prohibitive
\citep{Friedman1997, grossman2004learning}. 

To ease the computational burden, many works suggested various simplifying assumptions on the underlying distribution. 
One popular approach is to restrict attention to specific parametric forms such as the multivariate Gaussian or other distributions, for example copulas \citep{Lauritzen1996,Elidan2012}. A different approach, 
most relevant to our work and reviewed 
in Section~\ref{sec:background}, is based on the assumption that the multivariate distribution of each class follows a tree structured BN \citep{Tan2011a, Friedman1997}. In this case, the tree of each class may be learned by the computationally efficient algorithm of \cite{Chow1968} or extensions thereof \citep{Lafferty2012}. 
As we show in Section \ref{sec:method}, the implication of the tree assumption is that the likelihood ratio statistic of Eq. (\ref{eq:LR}) 
takes the form of a sparse linear combination of log-transformed univariate and bivariate density terms.
Hence, these tree based approaches  require only univariate and bivariate density estimates, and may be applied to high dimensional settings with a limited number of samples.  However, as real datasets may not follow a tree distribution, the resulting tree-based classifier may not accurately capture the optimal decision boundary. 

Motivated by the success of forest-based approaches, in Section~\ref{sec:method} we present a new semi-parametric classifier, denoted
as the Sparse Log-Bivariate density classifier (\texttt{SLB}). As its name suggests, it is also based on univariate and bivariate densities.
However, instead of imposing hard tree constraints, we assume that the decision boundary can be well approximated by a 
{\em sparse} linear combination of the logarithms of the univariate and bivariate densities.
Concretely, the \texttt{SLB} classifier is constructed as follows: first, we run a feature selection procedure, removing pairs of variables that are nearly independent. 
Next, we non-parametrically estimate for each class, its univariate densities as well as the bivariate densities of those pairs of variables not excluded by the feature selection procedure. 
Finally, we construct a linear SVM  in the space of these log-transformed density estimates. 

Our method can be viewed as a generalization of tree based classifiers. As 
we prove in Lemma \ref{lem:forestLR}, 
when the conditional distributions of both classes are tree-structured BNs,  
the log-likelihood ratio is a {\em linear} function of the univariate and bivariate log-transformed densities. 
The vector of coefficients of the resulting linear classifier, whose entries correspond to these log-transformed densities, is {sparse}.
Furthermore, its non-zero entries are integers, determined by the tree structure. Our approach eliminates the tree structure assumption, thus relaxing the integer constraints on the coefficients. Instead, it learns a sparse linear classifier in the space of these log-transformed densities.
Our approach also extends the recent work of \cite{fan2016feature},  which considered a logarithmic transformation of only univariate densities to address the same problem of binary classification.

In Section \ref{sec:theory} we present some theoretical properties of our classifier, in particular its asymptotics as sample size tends to infinity. In Section \ref{sec:results} we compare our approach to several popular classifiers
on both synthetic and real datasets. 
As our empirical results show, when the underlying distribution is forest structured, the accuracy of our classifier is comparable to methods
that explicitly assume  a forest structure. However, when the underlying distributions are not forest structured, but instead 
follow more complicated BN models, 
our more flexible method often outperforms
the compared competitors. Furthermore, our experiments highlight the importance of incorporating bivariate features.
Finally, as we illustrate with several real datasets, \texttt{SLB} is competitive to popular classifiers.  
We conclude the paper in Section \ref{sec:disc} with a discussion and potential extensions.

%
%
\section{Background} \label{sec:background}

In Section \ref{sec:BN} we briefly review
BNs and their relation to our approach. 
In Section \ref{sec:hilberschmidt} we describe the Hilbert-Schmidt independence criterion, used in our feature selection step. 

\subsection{Bayesian Network Classifiers}\label{sec:BN}

BNs provide a general framework to represent and infer about high dimensional densities \citep{koller2009probabilistic}. A BN is described by two components: (i) a directed acyclic graph $\mathcal{G}$ whose
nodes correspond to the random variables $X_1,\ldots,X_d$. 
(ii) a set of conditional densities, $\{f(X_i|Par_i)\}$, where
$Par_i$ denotes the parents of variable $X_i$ in the graph $\mathcal G$. For a root node $X_i$, without parents, $Par_i = \emptyset$, $f(X_i|Par_i)$ is its marginal density. The structure of the graph $\mathcal{G}$ encodes the statistical dependencies 
among the $d$ variables $X_i$. Specifically, given
the values of its parents, each random variable $X_i$ is conditionally
independent of all other variables $X_j$ which are not its descendants \citep{Lauritzen1996};  The full density $p(\x)$ is the product of all of these conditional densities. 
It can be shown that this indeed defines a valid joint density, which satisfies the conditional
independence properties encoded by $\mathcal G$ as outlined above. 

One popular application of BNs is to construct classifiers using the corresponding likelihood ratio. Since in practice, neither the graph structure nor the conditional densities are known, an essential part in constructing a BN-based classifier is learning, for each class label, its graph $\mathcal G$ and corresponding conditional densities. This problem is known as \emph{structure learning} \citep{beretta2018learning}. 

Unfortunately, even with as few as twenty variables, learning a general real-valued graphical model beyond the simple Gaussian BN can be computationally impractical
\citep{Friedman1997, grossman2004learning}. To overcome this computational burden, a common approach is to resort to \emph{tree-structured} networks. Under this assumption, the probability density $p(\x)$ factorizes as
\begin{equation} 
\label{eq:treedens}
p_T({\bf x}) = p(x_{m_1})\prod_{i=2}^{d} p(x_{m_i}|x_{m_{j(i)}}),
\end{equation}
where $(m_1,\ldots,m_d)$ is a permutation of $\{1,\ldots,d\}$ and $j(i)\in\{1,\ldots,i-1\}$ is the parent of $i$ in the tree.  An important property resulting from the tree-structured assumption is that for each class, 
the corresponding $d$-dimensional density depends only on its univariate and bivariate marginals. 
Given $n$ samples, the structure learning task simplifies to estimating the optimal tree structure and its associated univariate and bivariate densities. 

For a discrete multivariate distribution, a tree structure may be estimated by
the computationally efficient algorithm of \citet{Chow1968}. If the underlying distribution follows a tree structure, then 
as sample size $n\to\infty$ the estimated tree is consistent \citep{Chow_Wagner}. 
\citet{Tan2011a} extended this approach to more sparse structures, by 
removing weak edges from the learned tree. The result is a forest, or union of disjoint trees, with the following factorization,
\begin{equation} \label{eq:forestdens}
p_F({\bf x}) = \prod_{(i,j)\in E_F}\frac{ p(x_i,x_j)}{ p(x_i) p(x_j)} \prod_{i\in V_F} p(x_i),
\end{equation}
where $E_F$ and $V_F$ are the edge and vertex sets of the forest. 
 \cite{Lafferty2012} extended the \citeauthor{Chow1968} approach to the case of multivariate continuous data by using kernel density estimates of the univariate and bivariate densities in Eq. \eqref{eq:treedens}. In their paper the authors presented conditions under which this approach is consistent.

Several classifiers have been proposed which are based on specific forest models. 
Perhaps the simplest one is the popular naive Bayes classifier \citep{duda1973pattern,langley1992analysis}. This classifier assumes that the class conditional joint density is simply the product distribution $p_y(\x) = \Pi_i p_y(x_i)$, whose corresponding graph has no edges.
A more sophisticated model was considered by \citet{Park2011}. They assumed that \emph{both} classes are distributed according to the same a Markov chain. 
Namely, there exists a permutation $\pi \in \text{Sym}(d)$ such that for all $\x$ and $y \in \{-1,+1\}$, 
$$ p_{y}(\x) = p_y(x_{\pi_1}) p_y(x_{\pi_2}|x_{\pi_1}) \ldots p_y(x_{\pi_d}|x_{\pi_{d-1}}).
$$
Finally, \citet{Friedman1997} introduced the tree-augmented Naive-Bayes model (\texttt{TAN}). Their approach learns a tree structure over the set of variables for each class, and then apply a likelihood ratio based classifier. 

A different approach that makes a tree assumption is discriminative. 
For example, \citet{Tan2010} proposed to learn trees that are specifically tailored for classification. Their procedure fits each tree distribution to the observed data from that class while simultaneously maximizing the distance between the two distributions. 
Another example of a discriminative approach is the work of \cite{MeshiEbanElidanGloberson2013}, where instead of fitting a tree to the observed data, the authors suggest to learn a tree structure that directly minimizes a classification loss function. 
\subsection{Hilbert-Schmidt Independence Criterion} \label{sec:hilberschmidt}

An important component in our semi-parametric approach is a feature selection step,
whereby we remove bivariate densities that correspond to (nearly) independent variables.
We perform this step using the 
Hilbert-Schmidt Independence Criterion (HSIC) 
\citep{gretton2005measuring,gretton2005kernel}. 
For completeness we briefly review this method, though for simplicity, restricting attention to the case of univariate continuous real-valued random variables.

Let $(Z,W)$ be a pair of real-valued random variables on a compact domain
$\Omega_z\times\Omega_w\subset\mathbb{R}^2$ with joint density $P_{zw}$
and marginals $P_z,P_w$. HSIC is a method to assess if $Z$ and $W$ are independent, i.e., if $P_{z,w}=P_z\times P_w$.
The basic idea behind it, is that while $\Cov(Z,W)=0$ does not imply that $Z$ and $W$ are independent, having $\Cov(s(Z), t(W))=0$ for all bounded continuous functions $s$ and $t$ does actually imply independence. Unfortunately, going over all bounded continuous functions is not  
tractable. Instead, \cite{gretton2004behaviour} proposed evaluating $\sup_{ s \in \cF, \hspace{0.02in}t \in \cG} Cov(s(Z), t(W))$, where $\cF, \cG$ are universal Reproducing Kernel Hilbert Spaces (RKHS). They proved that if the RKHS are universal and the supremum is zero, then $Z$ and $W$ are independent. Next, in \cite{gretton2005measuring}, the authors introduced a quantity $\mbox{HSIC}(Z,W, \cF, \cG)$, which upper bounds the above supremum, and still satisfies that 
$\mbox{HSIC}(Z,W, \cF, \cG)=0$ if and only if $Z$ and $W$ are independent.
  
Defining $\mbox{HSIC}(Z,W, \cF, \cG)$ in its most generality requires some background in the theory of RKHS. Here, however, we only 
present the important properties that are essential for the fluent reading of this paper.
In particular, any RKHS $\cF$ is associated with a kernel $k(\cdot)$ and a mapping function $\phi$ from $\mathbb{R}$ to $\cF$, such that $k(x_1, x_2) = \langle \phi(x_1),\phi(x_2)\rangle_{\cF}$. Next, let $\cF, \hspace{0.03in} \cG$ be two RKHS with associated kernels $k(\cdot), \hspace{0.03in} l(\cdot)$ respectively. Let $(Z', W')$ be an independent copy of $(Z, W)$ with identical joint distribution. \cite{gretton2005measuring} showed that $\mbox{HSIC}(Z,W, \cF, \cG)$, defined as the Hilbert-Schmidt norm of the cross covariance operator of $Z$ and $W$,  can be equivalently expressed in terms of the two kernels as follows, assuming all expectations exist:
\begin{align*} 
   \mbox{HSIC}(Z,W, \cF, \cG)
    = &\ 
    \mathbb{E}_{ZWZ'W'}[k(Z, Z')l(W, W')]\\
    &+ \mathbb{E}_{ZZ'}[k(Z,Z')] \cdot \mathbb{E}_{WW'}[l(W,W')] \nonumber\\
    &-2\mathbb{E}_{ZW} \Big[ \mathbb{E}_{Z'}[k(Z, Z')]\mathbb{E}_{W'}[l(W, W')]\Big] \nonumber.
\end{align*}

Given a sample of $n$ independent pairs $S \equiv \{(z_i,w_i)\}_{i=1}^n$ drawn from the joint distribution $P_{zw}$,
one can estimate the HSIC by the following formula with complexity $O(n^2)$: 
\begin{align}
        \widehat{\mbox{HSIC}}(Z,W \cF, \cG)  = (n-1)-2\, \mbox{Tr}(KLHL), 
    \nonumber
\end{align}
where $H, K, L \in  \mathbb{R}^{n\times n}, K_{ij} \equiv k(z_i, z_j ), L_{ij} \equiv l(w_i, w_j ), H \equiv I_{n \times n} - (n-1)\bf{1}\bf{1}^{T}$, with $\bf{1}$ denoting a vector of all ones, and $\mbox{Tr}(A)$ denotes the trace of a matrix $A$.
Furthermore, as proven in \cite{gretton2005measuring}, this sample estimator is a nearly unbiased,  
\[
	\mathbb{E}_{S}[\widehat{\mbox{HSIC}}(Z,W, \cF, \cG) ] = \mathrm{HSIC}(Z,W, \cF, \cG)  + O(n^{-1}),
\]
with $\mathbb{E}_S$ denoting expectation over the sample $S$.

\section{The Sparse Log-Bivariate Density Classifier} 
\label{sec:method}

To motivate the construction of the sparse log-bivariate classifier, let us first consider the optimal log-likelihood ratio based classifier, when the distributions of both classes $y \in \{-1, +1\}$ are forest-structured, possibly with different graphs $G_y = (V_y, E_y)$. 
This is described in the following lemma, which follows direclty from Eq. (\ref{eq:forestdens}). 
\begin{lemma}\label{lem:forestLR}
If  both classes  are forest-structured 
then their log likelihood ratio is
\begin{align*} 
\log \Bigg( \frac{{p}_{+1}(\mathbf x)}{{p}_{-1}(\mathbf x)} \Bigg)
= & \sum_{(i,j)\in E_{+1}} \log {p}_{+1}(x_i,x_j)
+ \sum_{i\in V_{+1}} (1 - \text{deg}_{+1}(i)) \cdot \log {p}_{+1}(x_{i}) \\
& - \sum_{(i,j)\in E_{-1}} \log {p}_{-1}(x_i,x_j)
- \sum_{i\in V_{-1}} (1 - \text{deg}_{-1}(i)) \cdot \log {p}_{-1}(x_{i}), \nonumber
\end{align*}
where $\text{deg}_{y}(i)$ is the degree of $x_{i}$ in the graph of class $y$.
\end{lemma}

An immediate corollary of Lemma~\ref{lem:forestLR} is that if both classes follow a forest-structured graphical model then the optimal decision boundary is a 
{\em linear} combination of their univariate and bivariate log-transformed densities. Specifically, let $T_o:\mathbb{R}^d\to\mathbb{R}^{d(d+1)}$ be the oracle transformation that maps an input vector $\x \in \mathbb{R}^d$ to a vector containing all of $\x$'s univariate and bivariate log density terms for both classes,
\begin{align*} 
    T_o(\x) := \left\{ \log p_y(x_i, x_j)\ |\ y \in \{-1, +1\}\,\ i,j \in \{1, \ldots, d\} \right\},
\end{align*}
where $p_y(x_i,x_i)$ is an alias for $p_y(x_i)$.
In this notation, Lemma \ref{lem:forestLR} can be restated as
\begin{align*} 
    \log \text{LR}(\x) =  \w_o\trans T_o(\x),
\end{align*}
where the coefficients of $\w_o$ depend on the graph structures of both classes.
It follows that the Bayes classifier has the form
\begin{align} \label{eq:bayes_classifier}
    \hat{y}(\x) = \text{sign}(\w_o\trans T_o(\x) - b_o).
\end{align}
Under the forest model, the coefficients in $\w_o$ that correspond to the bivariate densities are all either $0,+1$ or $-1$, depending on the existence of edges in the respective graphical models.
Similarly, the coefficients that multiply the univariate densities are integers that depend on the degrees of the corresponding variables in the graphs.
The intercept $b_o$ depends on the class imbalance and misclassification costs.
Note that by the forest assumption, there are only $O(d)$ non-zero coefficients in the vector
$\w_o$. As its length is $O(d^2)$, it is thus {\em sparse}.

To derive the \texttt{SLB} classifier, we relax the forest-structure constraints, and simply assume that the optimal decision boundary can be well approximated by a sparse linear combination  of the univariate and bivariate log-densities,
\[
    \w\trans T_o(\x) -  b.
\]
To construct such a sparse vector $\w$, we first exclude bivariate densities that correspond to nearly independent pairs of variables. Specifically, we measure 
the strength of dependence between any pair of variables using their empirical HSIC measure, as defined in Section \ref{sec:hilberschmidt}, and filter out bivariate densities with estimated HSIC
values smaller than a given threshold $\lambda$.
We then fit a linear SVM to learn the coefficients of all the univariate log-densities and bivariate log-densities that correspond to the remaining pairs. Since $T_o$ is typically unknown, we replace it by a vector of estimated densities:
\begin{align} \label{eq:T_hat}
    \widehat T(\x) := \left( \log \widehat p_y(x_i, x_j) \right)_{y \in \{-1, +1\}\,\ i,j \in \{1, \ldots, d\}}
\end{align}

\begin{algorithm}[t]
    \caption{The Sparse Log-Bivariate Density Classifier (SLB)}
    \label{alg:SLB}
    \begin{algorithmic}
        \Statex
            {\bf Input:} Sample $(\x_1, y_1), \ldots, (\x_n, y_n)$ of feature vectors $\x \in \mathbb{R}^d$ and labels $y \in \{-1,+1\}$.\\
        \Statex {\bf Step 1:}  For each class $y \in \{+1, -1\}$, estimate statistical dependencies $\widehat{\mbox{HSIC}}$ between all pairs of variables $i,j \in \{1, \ldots, d\}$ and $y \in \{-1, +1\}$ and filter out weakly-dependent pairs of variables.
        \Statex{\bf Step 3:}
                Compute all univariate density estimates $\widehat{p}_y(x_i)$, and the bivariate density estimates $\widehat{p}_y(x_i,x_j)$
                for the variable pairs that were not filtered in the previous step.
      \Statex{\bf Step 4:}
       Fit a linear SVM $(\widehat \w, \widehat b)$ to the transformed samples $\{(\widehat{T}(\x_i), y_i )\}$, where $\widehat{T}$ is the log-density transformation defined in Eq. \eqref{eq:T_hat}.\\
        \Statex{\bf Output:}
        Classifier given by $\x \mapsto \text{sign}(\widehat\w\trans\widehat T(\x) - \widehat b)$.    
    \end{algorithmic}
\end{algorithm}

Lastly, setting $\lambda$ in a principled manner can be done in several ways. One approach is to run a cross-validation procedure and choose the threshold whose corresponding classifier has the smallest misclassification error. Another option  is to use a multiple hypothesis testing procedure on all pairwise $\widehat{\mbox{HSIC}}$ values. The \texttt{SLB} classifier is outlined in Algorithm~\ref{alg:SLB}. 

\section{ Theoretical Analysis} \label{sec:theory}
In this section we study the theoretical properties of the linear SVM solution $(\widehat \w, \widehat b)$ based on the estimated densities $\widehat p_{y}(x_i,x_j)$
given a training set of $n$ i.i.d. samples $D=\{(\x_{\ell}, y_{\ell})\}_{\ell=1}^{n}$. 
Concretely we prove that under certain conditions, the risk of the linear SVM solution $(\widehat \w, \widehat b)$ based on the estimated densities $\widehat p_{y}(x_i,x_j)$ converges to that of the optimal SVM classifier based on the exact densities $p_{y}(x_i,x_j)$.

To simplify the theoretical analysis we make the following two assumptions: (i) the input to the SVM classifier consists of all the univariate and bivariate log-transformed densities, without the HSIC-based filtering step; (ii) 
the $n$ training samples were split into two disjoint sets $D_0$ and $D_1$ of sizes $n_0$ and $n_1$, respectively, with $n_0+n_1=n$. 
The set $D_0$ is used to estimate the densities $\widehat p_{y}(x_i)$ and $\widehat p_{y}(x_i,x_j)$  for each class $y \in \{-1, +1\}$ and the set $D_1$ to construct the SVM classifier. 
Specifically, given the estimated densities, we apply the feature map  $\widehat T:\mathbb{R}^d\to\mathbb{R}^{d(d+1)}$ of Eq. \eqref{eq:T_hat}  to the $n_1$ samples in $D_1$ and construct a linear classifier using the transformed samples $\{(\widehat T(\x_{\ell}),y_{\ell})\}_{\ell=n_0+1}^{n}$. 
We add an additional feature equal to one to the feature map $T$, so that the intercept $b$ can be subsumed into $\w$.
The class label predicted  by a linear classifier $\w$ applied to a feature-mapped sample is thus sign$(\w\trans T(\x))$.

In the rest of this section, we define several risk functions and then formulate \texttt{SLB} as  an empirical risk minimizer.
The classification error rate, or 0-1 risk, of the classifier $(\w, T)$ is
\[
    R_{01}(\w,T) := \E_{(\x,y) \sim \mathcal D} \left[ \mathbbm{1} (y \neq \text{sign}(\w\trans T(\x))) \right].
\]
From Eq. \eqref{eq:bayes_classifier} it follows that $\w_o$ minimizes $R_{01}(\w, T_o)$.
The empirical 0-1 risk is the average loss over the $n_1$ samples in $D_1$,
\[
    \widehat{R}_{01}(\w,T) := \frac{1}{n_1} \sum_{\ell=n_0+1}^{n_1}  \left[ \mathbbm{1}(y_\ell \neq \text{sign}(\w\trans T(\x_\ell))) \right].
\]
Minimizing this risk with respect to $\w$ is computationally difficult \citep{BenDavidEironLong2000}.
To circumvent this difficulty, one may  consider instead  the risk and empirical risk with respect to the hinge loss $\phi(z) := \max\{0, 1-z\}$,
\[
    R_\phi(\w, T) := \E_{(\x,y) \sim \mathcal D} \left[ \phi(y\w\trans T(\x)) \right],
    \qquad
    \widehat{R}_\phi(\w, T) := \frac{1}{n_1} \sum_{\ell=n_0+1}^{n_1} \phi(y_\ell \w\trans T(\x_\ell)).
\]
The hinge loss is a convex function and therefore can be minimized efficiently.
Since the hinge-loss bounds  the 0-1 loss from above, minimizing the hinge loss is a way to control for the misclassification error.
A common formulation of the SVM classifier minimizes the empirical hinge loss with  an additional Tikhonov regularization term.
For the transformed data set $\{(\widehat T(\x_{\ell}),y_{\ell})\}_{\ell=n_0+1}^{n}$ this is 
\begin{align} \label{eq:soft_svm_tikhonov}
    \widehat \w_{\text{Tikhonov}} := \argmin_{\w} \left(  \widehat R_{\phi}(\w,\widehat T) + \lambda \| \w \|^2  \right),
\end{align}
where $\lambda > 0$ controls the regularization strength (see for example Section 15.2 of \citet{BenDavidShalevShwartz2014}).
For \texttt{SLB}, we consider instead the SVM solution  with Ivanov regularization,
\begin{equation} \label{eq:what} 
    \widehat \w := \argmin_{\w : \|\w\|_2 \le B}  \widehat R_{\phi}(\w,\widehat T),
\end{equation}
where the radius $B$ controls the complexity of the hypothesis class.
Both the Tikhonov form in Eq. \eqref{eq:soft_svm_tikhonov} and the Ivanov form in Eq. \eqref{eq:what} share the same regularization path \citep{Oneto2015}.
Namely, for every choice of Tikhonov regularization parameter $\lambda$
there is an Ivanov regularizer $B_\lambda$ which yields the same solution.
Finally, we denote by $\w_\phi$ the  SVM population minimizer using the oracle feature map $T_o$ 
\begin{equation}
\label{eq:wstar}
        \w_\phi := \argmin_{{\w: \|\w\|_2 \le B}}  R_\phi(\w,T_o).
\end{equation}
In the rest of this section, we prove that as the sample size tends to infinity, the risk of the \texttt{SLB}  classifier $R_\phi(\widehat \w, \widehat T)$
converges to the oracle risk $R_\phi(\w_\phi,  T_o)$ and give high probability bounds on their difference.

\subsection{Convergence of the Data-dependent Feature Map}

We begin by studying  the convergence rate of the estimated feature map $\widehat{T}$ to the oracle map $T_o$. 
For simplicity, we use the bivariate notation $\widehat{p}_y(x_i, x_i)$ to also denote  univariate density estimates $\widehat{p}_{y}(x_i)$. 
For our theoretical analysis we need 3 assumptions on the class conditional densities and their estimation procedures:
\begin{assumption} \label{assumption:density_bound}
    There are constants $p_{\min}, p_{\max} > 0$ such that for all $\x \in \Omega$
    and all  $y,i,j$ we have that \( p_{\min} \le p_y(x_i, x_j) \le p_{\max} \).
\end{assumption}

\begin{assumption} \label{assumption:kde_uniform_convergence}
    There exists a rate decay function $U(n_0)$ that satisfies $U(n_0) \xrightarrow{n_0 \to \infty} 0$
    such that for every $y,i,j$, the following bound holds with high probability, 
    \begin{align*} 
        \sup_{\x \in \Omega}  | \widehat{p}_y(x_i, x_j) - p_y(x_i, x_j) | < U(n_0).
    \end{align*}
\end{assumption}
\begin{assumption} \label{assumption:finite_expectation}
    The following expectation is finite,
    \begin{align*} 
        E(n_0,d)
        :=        
        \E_{\x}
\|\widehat{T}(\x) - T_o(\x)\|_2
        =
        \E_{\x}
        \sqrt{
            \textstyle \sum_{y,i,j} (\log \widehat p_y(x_i,x_j) - \log p_y(x_i,x_j))^2
        }        
    \end{align*}
    and $E(n_0,d) \xrightarrow{n_0 \to \infty} 0$.
\end{assumption}
\begin{rmrk}
Assumption \ref{assumption:kde_uniform_convergence} holds under various conditions on the underlying
density and method of estimation.
For $\beta-$H\"older bivariate densities, the minimax bound $U(n) = n^{-\beta/(2\beta+2)}$
holds (up to log factors) with high probability, both for kernel density estimation \citep{Liu2011, Jiang2017}
and for k-nearest-neighbor density estimation \citep{DasguptaKpotufe2014}.
\end{rmrk}

\begin{rmrk}
A potential problem with non-parametric kernel density estimators is that $\widehat p_y(x_i,x_j)$ may be zero (or even negative) at some values of $(x_i,x_j)$.
Then, the $\log$ term either explodes or is underfined. This problem may be avoided  
by constructing an adjusted density estimator that satisfies $C_{\max} \geq \widehat p_y(x_i,x_j) \ge c_{\min}$ with say  $c_{\min} = p_{\min}/2$
and $C_{\max}=2 p_{\max}$.
It can be shown that in this case Assumption \ref{assumption:finite_expectation} will hold. 
\end{rmrk}

\begin{rmrk}
The assumption that $p_{\min} >0$ may be relaxed to an assumption that all $d$ coordinates of $X=(X_1,\ldots,X_d)$ are sub-exponential random variables. 
In this case, for a suitable threshold $\epsilon$, one can separate the possibly infinite support of the distribution to a compact domain  $\Omega_\epsilon=\{{\bf x}\,|\, p({\bf x}|y) >\ \epsilon\}$, on which our theoretical analysis holds. On the remaining part the SVM risk (average of the hinge loss) can be bounded by $C\Pr[{\bf x}\not\in\Omega_\epsilon]$, where the constant \(C\) depends on the rate of decay of the density. 
\end{rmrk}

The following lemma, proved in the appendix quantifies the convergence rate of
the feature map $\widehat T$ to the oracle map.  
\begin{lemma} 
    \label{lem:consistencyT}
    Under Assumption \ref{assumption:density_bound},
    \begin{align*}
        &\| T_o(\x) \|_2
        <
        \sqrt{d(d+1)} L,
        \quad \text{where} \quad
        L: = \max \{ |\log p_{\min}|, |\log p_{\max}| \}.
    \end{align*}
    Under Assumptions \ref{assumption:density_bound} and \ref{assumption:kde_uniform_convergence}, the following bound holds w.h.p.    
    \begin{align*}
        &\sup_{\x \in \Omega} \|\widehat{T}(\x) - T_o(\x)\|_2 < \frac{2}{p_{\min}} \sqrt{d(d+1)} U(n_0).
    \end{align*}
\end{lemma}
\begin{corollary} \label{cor:sup_bound_That}
    Under Assumptions \ref{assumption:density_bound} and \ref{assumption:kde_uniform_convergence}, w.h.p.    
    \begin{align*}
        \sup_{\x \in \Omega} \|\widehat T(\x)\|_2 < \sqrt{d(d+1)}\left(L+\frac{2}{p_{\min}} U(n_0)\right).
    \end{align*}
\end{corollary}

\subsection{Convergence of the SVM Risk to the Oracle Risk}
In this section we present our main theoretical result.
The proofs appear in the appendix.
We begin with a technical lemma that bounds the difference in  risks
 of using the exact log densities $T_o$ compared to using the estimated log densities $\widehat{T}$.
\begin{lemma} \label{lem:empirical_risk_estimated_T_vs_oracle_T}
    Under Assumption \ref{assumption:finite_expectation},
    for any $\w \in \mathbb{R}^{d(d+1)}$,
    \begin{align*}
        | R_\phi(\w, \widehat{T}) - R_\phi(\w, T_o)|
        \le 
        \| \w \|_2E(n_0,d).
    \end{align*}
    This holds for the empirical risk as well,
    \begin{align*}
        | \widehat{R}_\phi(\w, \widehat{T}) - \widehat{R}_\phi(\w, T_o)|
        \le 
        \| \w \|_2E(n_0,d).
    \end{align*}

\end{lemma}
We now show that for the hinge loss, the population risk of the estimated 
$\widehat \w$ of Eq. \eqref{eq:what}, converges in probability to the population risk of the oracle SVM solution $\w_\phi$, defined in Eq. \eqref{eq:wstar}. In other words, $R_{\phi}(\widehat \w, \widehat T)$ converges to $R_{\phi}(\w_\phi, T_o)$ in probability:
\begin{theorem} \label{thm:convergence_to_oracle_risk}
    Let $\widehat{T}$ be the feature map  constructed using $n_0$ samples
    and let $\widehat{\w}$ be the SVM classifier of \texttt{SLB}, trained using $n_1$ samples.
    Then, under Assumptions \ref{assumption:density_bound},\ref{assumption:kde_uniform_convergence} and \ref{assumption:finite_expectation}, the following bound holds w.h.p.    
    \[
        R_\phi(\widehat{\w},\widehat{T}) - R_\phi(\w_\phi, T_o)
        <
        B
        \left(
            E(n_0,d)
            +
            \sqrt{
                d(d+1)
                \tfrac{\ln n_1}{n_1}
            }
            \left(
                2L
                +
                \tfrac{2}{p_{\min}}
                U(n_0)
            \right)
        \right).
    \]
\end{theorem}
The terms $E(n_0,d)$ and $U(n_0)$ are due to errors in the univariate and bivariate density estimates,
whereas the term $\sqrt{\ln  n_1/ n_1}$ is an SVM generalization term.

\section{Experimental Results} 
\label{sec:results}


We compare the predictive accuracy of \texttt{SLB} to several widely used classifiers both on synthetic data and on various public
datasets.
 Our method  may be tuned to improve performance on a specific domain by considering such things as different kernels and bandwidths in the density estimation step, as well as different misclassification costs and penalty terms for the
norm of the vector of coefficients ($\lambda$) in the construction of the SVM classifier. However, we refrain from doing so and instead demonstrate SLB's competitiveness across a broad range of datasets. 
The seven classification methods we tested appear below. All compared methods were implemented in the \texttt{R} programming language. When required, univariate or bivariate densities were estimated by the \texttt{ks} package.  
\begin{itemize}
   \item  {\bf SLB}: Our sparse log-bivariate density classifier. We estimated all pairwise HSIC values using the \texttt{dHSIC} package with a Gaussian kernel. The SVM classifier was constructed by the \texttt{e1071} package with a default parameter $\lambda=1/2$. 
     \item	{\bf LU}:  Log-univariate density classifier. This classifier trains an SVM model using only the log univariate densities of the $d$ variables of each class, as its input features. This is similar to the approach of \cite{fan2016feature}, that trained a logistic regression model rather than an SVM, on the same features. 
     \item   {\bf 5NN}: 5 nearest-neighbor classifier, as implemented in the \texttt{FNN} package.
     \item   {\bf SVM}:  Support vector machine classifier with the radial basis function kernel, as implemented in the \texttt{e1071} package. Note that by default $\lambda=\frac{1}{2}$.
      \item  {\bf NB}: Naive Bayes classifier. Here we used our own implementation.
     \item   {\bf TAN}: Tree-augmented naive Bayes. We used the \texttt{mpmi} package to estimate the mutual information between each pair of variables, and the \texttt{optrees} package to estimate the maximum spanning tree of each class.
     \item   {\bf RF}: Random forest classifier as implemented in the \texttt{randomForest} package, with 50 trees and the default number of $\sqrt{d}$ randomly selected variables at each split.
    \end{itemize}

\subsection{Synthetic Data Experiments} 
\subsubsection{Forest Structured BNs}
We begin with the simple setting where the underlying distribution of each class is a forest structured BN with $d=20$ variables. We consider the following configurations that differ in number of training samples, class imbalance proportion and complexity of the Conditional Probability Distributions (CPDs) that parametrize the BNs. Concretely, our simulations follow a $5\times 2^{2}$ factorial design:
\begin{itemize}
\item
{\bf Sample size of the training data:} $n=200, 400, 600, 800, 1000$.
\item 
{\bf Class imbalance:} either balanced 50\%-50\% or imbalanced 75\%-25\%.
\item
{\bf Complexity of the Conditional Probability Distributions:} Let $Z$ be the single parent of $X$ in the forest. We consider the following two cases for $P(X|{\bf Z}=z)$: \newline (i) a simple setting where $X|{\bf Z}=z \sim N(z,1)$. \newline (ii) a more complex setting in which 
\begin{equation}\nonumber
  X|{\bf Z}=z \sim
    \begin{cases}
      t+z & \text{w.p. $\frac{1}{2}$}\\
       \frac{1}{2}N(z+1,1)+\frac{1}{2}N(z-1,1) & \text{w.p. $\frac{1}{2}$}
     \end{cases}   ,    
\end{equation}
where $t$ is a student-t distribution with 5 degrees of freedom. \newline In both cases, root variables follow a standard Gaussian distribution.
\end{itemize}

For each combination of factor levels, $100$ BNs were generated. A training dataset drawn from each model was used to construct the different classifiers. Finally, we evaluate the predictive accuracy on an independent test set of $1000$ i.i.d. samples generated from the same model. In each test set, half of the observations were generated from each class. Average misclassification error rates from the average $100$ replicates are shown in Appendix ~\ref{simulations}, 
Figures ~\ref{forest_BNs_Gaussian_balanced}--\ref{forest_BNs_complex_imbalanced}. The full results appear in Appendix ~\ref{simulations}, Table ~\ref{fig:forest_Simulation_results}. In addition, in each setting, we compare the classifier with the best performance to each of the other classifiers, using a two-sided Wilcoxon signed-rank test. Since there are 6 comparisons per configuration setting, we use Bonferroni-correction with $\alpha=0.05/6$. The classifier with the best performance, as well as classifiers that were found to be statistically compatible (i.e., the null hypothesis is not rejected) are printed in the table in boldface. As shown, under the Gaussian setting, \texttt{TAN} either achieves the minimum error rate or it is comparable to the best performing classifier. This is somewhat to be expected since the \texttt{TAN} classifier is based on the Chow-Liu algorithm \citep{Chow1968}, that is consistent under the \emph{Gaussian-tree} setting. In our \emph{Gaussian-forest} setting, we expect the tree constructed by \texttt{TAN} to include all true forest edges, and a few additional spurious  ones that make the forest a connected graph. Nevertheless, in all the $5\times 2^2$ configurations, as the sample size increases, the gap between \texttt{SLB} and \texttt{TAN} decreases. 
It is also apparent that both \texttt{SLB} and \texttt{TAN} models outperform all other classifiers under the Gaussian-imbalanced configuration. 

\subsubsection{General BNs}
Next we consider a more challenging setting where each class follows a BN in which all non-root nodes have three parents. For a r.v. $X$, at a non-root node, let $Par_X \equiv \{Z_1, Z_2, Z_3\}$ be its three parents. We consider the following two settings for $P( X| Par_X)$: 
\newline (i) a simple setting where $X|{\bf Par_X }\sim N(\sum_{Z_i \in {\bf Par_X}}z_i, 1)$ 
\newline(ii) a more complex setting in which
\begin{equation}\nonumber
  X|{\bf Par_X} \sim
    \begin{cases}
      t+\sum_{Z_i \in {\bf Par_X}}z_i & \text{w.p. $\frac{1}{2}$}\\
       \frac{1}{2}N(z_1+z_2,1)+\frac{1}{2}N(z_2+z_3,1) & \text{w.p. $\frac{1}{2}$}
     \end{cases}   ,    
\end{equation}

where $t$ is a student-t distribution with 5 degrees of freedom. \newline As before, we set $d=20$ and root variables follow a standard Gaussian distribution. 
To accommodate further variety, we extend our factorial design, and allow for partial common structure between the BNs of the two classes; In the simpler setting, the two networks are constructed independently of each other. In the more challenging setting, roughly a third of the nodes share the same parent sets and associated 
conditional probability distributions. Overall our simulations now follow a $5\times 2^3$ factorial design. We repeat the same protocol as in the forest setting, with 100 BNs generated for each configuration. Classifiers are evaluated on an independent test set of 1000 samples. As  before, in these test sets, half of the observations were generated from each class. 

To highlight the importance of removing bivariate densities corresponding to nearly independent variables, we also consider a variant of \texttt{SLB} in which we skip the first step of Algorithm \ref{alg:SLB}. The resulting classifier denoted \texttt{SLB-}, constructs a linear SVM using {\em all} log bivariate densities. 

Average misclassification error rates from the $100$ replicates are shown in Appendix ~\ref{simulations}, Figures ~\ref{general_BN_Gaussian}--\ref{general_BN_complex}. The full results, including standard deviations, are displayed in  Appendix ~\ref{simulations}, Table ~\ref{fig:general_BNs_Simulation}. Again, we compare the classifier with the best performance to each of the other classifiers, using a two-sided Wilcoxon signed-rank test, at the Bonferroni-corrected $\alpha=0.05/7$ level (since now we have a total of 8 classifiers). The classifier with the best performance, as well as classifiers that were found to be statistically compatible are printed in boldface.

As can be seen, in all configurations, \texttt{SLB} achieves the lowest misclassification error rate, whereas \texttt{TAN} attains compatible results in only five (out of 40) configurations, all five having Gaussian distributions.  Lastly, \texttt{SLB} is significantly more accurate than \texttt{SLB-}, particularly at small sample sizes. As sample size increases, the estimated bivariate densities of independent variables become closer to the product of the univariate densities. In this case their logarithm is approximately the sum of the log of the corresponding univariate densities. As the latter are already incorporated into the SVM model, the effect of removing such pairwise densities is negligible, and \texttt{SLB} and \texttt{SLB-} achieve similar error rate. 
%
\subsection{Real Data Experiments}
\begin{table}[t]
\scalebox{0.57}{
\begin{tabular}{c | ccc |c c c c c c c} \hline  \hline
Dataset       &$d$ & $n$ & $(n_1, n_2)$ & SLB & LU &  TAN & NB & RF &SVM (RBF) & 5NN \\ \hline

Blood & $4$ & $748$ & $(570,178)$ & ${\bf 40 \pm 6.3 }$ & $ 49.7 \pm 1.7 $ & $ 42.6 \pm 2.9 $ & ${\bf 36.8 \pm 6.7 }$ & ${\bf 40.1 \pm 3.7 }$ & $ 44.1 \pm 3.2 $ & ${\bf 37.4 \pm 3.9 }$ \\

Climate Model Simulation Crashes & $18$ & $540$ & $(46,494)$ & ${\bf 25.2 \pm 9.8 }$ & $ 50  \pm 1.3$ & $ 43.8 \pm 2.7 $ & $ 47.8 \pm 3 $ & $ 46.7 \pm 4.9$ & $ 42.4 \pm 4.8 $ & ${\bf 36.2 \pm 7.9}$ \\

Hill-valley without noise & $100$ & $606$ & $(305,301)$ & ${\bf 38.9 \pm 7.1 }$ & 	$ 51.3 \pm 6.1 $ 	& 	$ 46.7 \pm 3$ 	&	 $ 43.9 \pm 3.1$	 &	 $ 46.6 \pm 3.1 $	 &	 $ 47.2 \pm 2.2 $	&	 $ 47.2 \pm 6.6 $ \\

Hill-valley with noise & $100$ & $606$ & $(307,299)$ & ${\bf 37.7 \pm 2.3 }$ & 	$ 51.8 \pm 6.4$ 	& 	$ 47.3 \pm 61 $ & 	$ 44.3 \pm 4.6$ 	& 	$ 47.9 \pm 3.6 $ 	& 	$ 47 \pm 5.8 $ 		& 	$ 48.4 \pm 2.4 $ \\

Ionosphere & $32$ & $351$ & $(126,225)$ & ${\bf 7.5 \pm 5 }$ 		      & 	$ 45.4 \pm 5.4 $ 	& 	${\bf 7.6 \pm 4.1 }$ & $ 10.2 \pm 4.3 $ & ${\bf 7.9 \pm 4.4 }$ & ${\bf 8 \pm 4.3 }$ & $ 22.1 \pm 7.4 $\\ 

Liver & $6$ & $345$ & $(145,200)$ & ${\bf 30.8 \pm 2.7 }$ & $ 48.6 \pm 2.8 $ & $ 35.5 \pm 5.2 $ & $ 38.1 \pm 4.08 $ & ${\bf 29.9 \pm 7.3 }$ & ${\bf 33.2 \pm 4.7 }$ & $ 40.6 \pm 3.7 $ \\

Ozone & $71$ & $1847$ & $(1719,128)$ & $ 32.3 \pm 2.8 $ & $ 48.8 \pm 1.1 $ & $ 42.5 \pm 3.9 $ & ${\bf 24.9 \pm 3.3 }$ & $ 40.1 \pm 5.1$ & $ 49.6 \pm 0.8$ & $ 41.4 \pm 3.7$ \\

Parkinson & $22$ & $195$ & $(48,147)$ & $ 18.2 \pm 7.1 $ & $ 54.7 \pm 3.9 $ & $ 21.3 \pm 5.5 $ & $ 20.6 \pm 7.5 $ & ${\bf 13.1 \pm 1.7}$ & $ 27.6 \pm 6.7 $ & ${\bf 13.6 \pm 6.3}$ \\

Pima & $8$ & $768$ & $(500,268)$ & ${\bf 28.6 \pm 3.8 }$ & $ 49.2 \pm 1.5 $ & $ 34.5 \pm 6.6 $ & ${\bf 27.4 \pm 4.9 }$ & ${\bf 26.8 \pm 2.4}$ & $ 29 \pm 1.4 $ & $ 31.2 \pm 2.6 $ \\


Pulsar stars & $8$ & $17898$ & $(16259,1639)$ & ${\bf 7.1 \pm 1.3 }$ & $ 49.6 \pm 2.6 $ & $ 8.6 \pm 0.8$ & $ 8.31 \pm 0.9$ & $ 8.2\pm 1.2$ & $ 9.3 \pm 1.2$ & $ 8.8\pm 1.2$ \\

Ringnorm & $20$ & $7400$ & $(3664,3736)$ & ${\bf 1.4 \pm 0.3 }$ & $ 31.8 \pm 1.6 $ & $ 2.7\pm 0.9 $ & ${\bf 1.4\pm 0.3}$ & $ 3.8\pm 0.5$ & $ 1.4\pm 0.3$ & $ 31.7 \pm 0.8$\\ 

Sonar & $60$ & $208$ & $(111,97)$ & ${\bf 18.1 \pm 6.5 }$ & $ 52.9 \pm 5.1 $ & ${\bf 20.5 \pm 4.9 }$ & $ 26.2 \pm 9.2$ & ${\bf 19.6 \pm 9.8 }$ & ${\bf 20.1 \pm 5.9 }$ & $ 22.9 \pm 2.7$\\ 

Spectf & $44$ & $80$ & $(40,40)$ & ${\bf 21.2 \pm 7.1 }$ & $ 45 \pm 21.8 $ & ${\bf 26.2 \pm 20.4 }$ & ${\bf 22.5 \pm 7.1}$ & ${\bf 21.2 \pm 13.7 }$ & ${\bf 20 \pm 11.2 }$ & $ 27.5 \pm 14.4 $ \\

WI prog & $32$ & $198$ & $(151,47)$ & ${\bf 35.7 \pm 13.5 }$ &  $52.7 \pm 11.6$ & ${\bf 42.9 \pm 6.9 }$ & ${\bf 39.8 \pm 9.6}$ & $42 \pm 4.9$ & $44 \pm 4.1$ & ${\bf 40.3 \pm 6.3}$ \\

WI diag & $30$ & $569$ & $(357,212)$ & $ 4.52 \pm 2.2 $ & $ 47.9 \pm 4.7 $ & $ 11.5 \pm 2.1$ & $ 6.3\pm 3 $ & $ 4.8\pm 2.3$ & ${\bf 2.9 \pm 2.4 }$ & $ 4\pm 2.6$ \\

Vertebral & $6$ & $310$ & $(210,100)$ & $ 23.2 \pm 5.3 $ & $ 50 \pm 1.2 $ & ${\bf 18.8 \pm 4.9 }$ & $ 22.5 \pm 3.8 $ & ${\bf 18.8 \pm 7.3}$ & ${\bf 19.2 \pm 7.9}$ & ${\bf 22.1 \pm 7}$ \\
\hline \hline
\end{tabular}}
\caption{Summary of balanced error rates for the various classifiers evaluated on datasets from the UCI ML and Kaggle Repository.
$n$ is the total number of labeled samples, $n_1$ and $n_2$ are the number of samples in the `positive' and `negative' classes respectively. The classifier with the best performance, as well as classifiers that achieve comparable error rates are printed in boldface (see main text for details).}
\label{tab:dataresults}
\end{table}

We next evaluate the various classifiers on 16 real datasets, publicly available at the UCI Machine Learning and Kaggle\footnote{www.kaggle.com} Databases \citep{UCI}. The datasets we consider differ in sample size and dimensionality. In all datasets, instances with missing values were removed. In datasets where features are not commensurate, such as the Ozone dataset in which one feature is the temperature and another is the wind speed, they were standardized prior to applying 5NN and SVM (RBF), both of which are known to be sensitive to the scale of the data. 

The misclassification error was estimated by 5-fold cross-validation, with the folds  sampled in a stratified manner so that they have approximately the same proportions of class labels as the full dataset. 
For each dataset, the various classifiers were learned on the same training sets and their performance evaluated on the same test sets. 
Since some datasets had a large class imbalance, we evaluated each classifier by its Balanced Error Rate (BER),
\[
    \mbox{BER} = 1 - \frac{1}{2} (\mbox{Sensitivity} +\mbox{Specificity}).
\]
Sensitivity is the true positive rate, equal to the proportion of predicted positives from the positive set. Specificity is the true negative rate, equal to the proportion of predicted negatives
from the negative set. Note that the misclassification error rates for the simulated datasets in the previous subsection were estimated on a test set with an equal number of samples in each class, and hence are equivalent to the BER.
 
The empirical mean and standard deviation of the BER taken across the 5 cross-validation folds are given in Table \ref{tab:dataresults}. For each dataset, the classifier with the smallest error rate and classifiers that are not significantly different from it, according to a a two-sided Wilcoxon signed-rank test, appear in bold. Note that the tests were performed at Bonferroni-corrected $\alpha=0.05/6$ level.

As shown, in 13 out of 16 datasets, \texttt{SLB}  either achieves the minimum misclassification rate or is on par with the best performing model. Comparing 
\texttt{SLB} to \texttt{LU} and \texttt{NB} which only uses the univariate log density features, highlights the importance of including at least some of the bivariate features.
Finally, note that in 15 out of 16 datasets, \texttt{SLB} achieves balanced error rates that are either better or compatible to those of the \texttt{TAN} classifier. These results demonstrate the potential benefit of \texttt{SLB} over the \texttt{TAN} classifier that relies on a tree assumption, which may be violated in practice.

%
%
\section{Discussion}
\label{sec:disc}

In this work, we proposed the sparse log-bivariate density classifier (\texttt{SLB}), a semiparametric procedure that 
generalizes tree and forest based classification approaches. As in forest-based methods, \texttt{SLB}
requires the estimation of only univariate and bivariate densities. However, it is more flexible, and does not require learning the structure of the class conditional distributions. 
At test time, \texttt{SLB} needs to evaluate
the estimated log densities $\widehat{T}(\x)$ at new instances $\x$.
Using a standard kernel density estimator, this requires to have at test time the original training data. Nevertheless,  
since the computation involves only 1-dimensional and 2-dimensional density estimates, there are efficient methods to approximate these estimated densities
without access to the original data \citep{ParikshitLeeMarchGrey2009}.

\texttt{SLB} uses the HSIC measure to rank the bivariate densities and applies a  cross-validation procedure to select only some of them. 
A different approach is to directly incorporate a sparsity-inducing penalty term, into the SVM loss function, similar to
the approaches presented in \cite{neumann2005combined} and \cite{zhu20041}.

In recent years, the winners of many data modeling competitions
made use of extensive feature engineering, whereby many new features are generated from the original features in the dataset (for example, \cite{Koren2009, NarayananShiRubinstein2011}).
Our work suggests that the estimated log densities $\log \widehat p_y(x_i)$ and $\log \widehat p_y(x_i,x_j)$
are interesting features to consider for a wide range of classification problems.
Of course, one may choose to also keep the original features
as in \cite{fan2016feature} or to combine the log density features with other non-linear features.


\vspace{0.2in}
\acks{This work was supported in part by a Texas A\&M-Weizmann research grant from Paul and Tina Gardner.
BN is incumbent of the William Petschek professorial chair in mathematics. Part of this work was done while BN was on sabbatical at the Institute for Advanced Study at Princeton. BN would like to thank the IAS and the Charles Simonyi endowment for their generous support.}


%
%

\appendix
\section{Proofs}


\begin{proof}{\bf of Lemma \ref{lem:consistencyT}.}
The first part follows immediately from Assumption \ref{assumption:density_bound}, since
\begin{align*}
    \| T_o(\x) \|_2^2 = \sum_{y,i,j} \log^2 p_y(x_i,x_j) < d(d+1) \max\{ |\log p_{\min}|, |\log p_{\max}| \}^2.
\end{align*}
We now turn to the second part of the lemma.
\begin{align*}
    \sup_{\x \in \Omega}& \| \widehat T(\x) - T_o(\x)\|_2^2
    =
    \sup_{\x \in \Omega}
    \sum_{y,i,j} ( \log \widehat p_y(x_i,x_j) - \log p_y(x_i,x_j))^2.
\end{align*}
By the mean value theorem, for any $a,b>0$ there exists some $\xi \in [a,b]$ such that
\(
    \log b - \log a
    =
    (b-a)/\xi
\).
It follows that
\begin{align} \label{eq:log_diff_bound}
   \frac{|b-a|}{\max\{a,b\}}
    \le
    |\log b - \log a|
    \le
    \frac{|b-a|}{\min\{a,b\}}.
\end{align}
Hence,
\begin{align*}
    \sup_{\x \in \Omega}& \| \widehat T(\x) - T_o(\x)\|_2^2
    \le\sup_{\x \in \Omega}
    \sum_{y,i,j}
    \left(
        \frac
        {\widehat p_y(x_i,x_j) - p_y(x_i,x_j)}
        {\min\{\widehat p_y(x_i,x_j), p_y(x_i,x_j)\}}
    \right)^2.
\end{align*}
By Assumptions \ref{assumption:density_bound} and \ref{assumption:kde_uniform_convergence}, w.h.p. $\min\{\widehat p_y(x_i,x_j), p_y(x_i,x_j)\} > p_{\min}/2$, therefore
\begin{align*}
    \sup_{\x \in \Omega}& \| \widehat T(\x) - T_o(\x)\|_2^2\\
    &\le
    \frac{4}{p_{\min}^2}
    \sup_{\x \in \Omega}
    \sum_{y,i,j}
    \left(
        {\widehat p_y(x_i,x_j) - p_y(x_i,x_j)}
    \right)^2\\
    &\le
    \frac{4}{p_{\min}^2}
    \sum_{y,i,j}
    \sup_{\x \in \Omega}
    \left(
        {\widehat p_y(x_i,x_j) - p_y(x_i,x_j)}
    \right)^2\\
    &<
    \frac{4}{p_{\min}^2} d(d+1) U^2(n_0).    
\end{align*}

\end{proof}


\begin{proof}{\bf of Lemma \ref{lem:empirical_risk_estimated_T_vs_oracle_T}.}
By definition,
\[
    | R_\phi(\w, \widehat{T}) - R_\phi(\w, T_o)|
    =
    \Bigg| \E_{(\x,y) \sim \mathcal D} \left[ \phi(y \w\trans \widehat{T}(\x)) - \phi(y \w\trans T_o(\x) ) \right] \Bigg|.
\]
Using the triangle inequality we have
\begin{align*}
    | R_\phi(\w, \widehat{T}) - R_\phi(\w, T_o)|
    \le
    \E \big| \phi(y  \w\trans \widehat{T}(\x)) - \phi(y \w\trans T_o(\x))\big|.
\end{align*}
Since the hinge-loss $\phi$ is 1-Lipschitz and $|y|=1$, we can simplify the bound further to
\begin{align*}
    | R_\phi(\w, \widehat{T}) - R_\phi(\w, T_o)|
    \le
    \E \big| y  \w\trans ( \widehat{T}(\x) - T_o(\x)) \big|
    =
    \E \big| \w\trans ( \widehat{T}(\x) - T_o(\x)) \big|.
\end{align*}
By the Cauchy-Schwarz inequality and Assumption \ref{assumption:finite_expectation},
\begin{align*}
    | R_\phi(\w, \widehat{T}) - R_\phi(\w, T_o)|
    \le
    \| \w \|_2\E   \| \widehat{T}(\x) - T_o(\x) \|_2
    =
    \| \w \|_2 E(n_0,d)
    .
\end{align*}
The bound for the empirical risk is proved in the same manner.
\end{proof}

\begin{proof}{\bf of Theorem \ref{thm:convergence_to_oracle_risk}.}
    We decompose the difference of risks into 4 terms,
    \begin{align} \label{eq:generalization_bound_decomposition}
        R_\phi(\widehat{\w},\widehat{T}) &- R_\phi(\w_\phi, T_o)
        \\
        =&
        \left(R_\phi(\widehat{\w}, \widehat{T}) - \widehat{R}_\phi(\widehat{\w}, \widehat{T})\right)
        +
        \left(\widehat{R}_\phi(\widehat{\w}, \widehat{T})-\widehat{R}_\phi(\w_\phi, \widehat{T})\right) \nonumber
        \\
        &+\left(\widehat{R}_\phi(\w_\phi, \widehat{T})-\widehat{R}_\phi(\w_\phi, T_o)\right)
        +
        \left(\widehat{R}_\phi(\w_\phi, T_o) - R_\phi(\w_\phi, T_o)\right) \nonumber.
    \end{align}
    We  now bound each of these terms separately.
    To bound the first and fourth terms
    we  apply a generalization bound for the soft-margin SVM.
    First recall that by Lemma \ref{lem:consistencyT}, $\| T_o(\x) \|_2 < \sqrt{d(d+1)} L$.
    It follows from Theorem 26.12 of \cite{BenDavidShalevShwartz2014}, that the following bound is satisfied w.h.p. over $D_1 \sim \mathcal D^{n_1}$.
    \begin{align} \label{eq:bound_R_Toracle_minus_Rhat_Toracle}
        \sup_{\w: \|\w\| \le B} & |R_\phi(\w, T_o) - \widehat{R}_\phi(\w,T_o)|
        < \sqrt{d(d+1)}L B \sqrt{\frac{\ln n_1}{n_1}}.
    \end{align}
    In particular, this gives a high probability bound on the fourth term of Eq. \eqref{eq:generalization_bound_decomposition}.
    Using Corollary \ref{cor:sup_bound_That} we can similarly bound the first term of  Eq. \eqref{eq:generalization_bound_decomposition}.
    w.h.p over $D \sim \mathcal D^{n_0+n_1}$,
    \begin{align} \label{eq:bound_R_That_minus_Rhat_That}
        |R_\phi(\widehat\w, \widehat T) - \widehat{R}_\phi(\widehat \w, \widehat T)|
        < \sqrt{d(d+1)}\left(L+\frac{2}{p_{\min}} U(n_0)\right) B \sqrt{\frac{\ln n_1}{n_1}}.
    \end{align}
    To bound the second term of Eq. \eqref{eq:generalization_bound_decomposition}, note that  $\widehat{\w}$ is the minimizer of $\widehat{R}_\phi(\w, \widehat{T})$, hence
    \begin{align*}
        \widehat{R}(\widehat{\w}, \widehat{T})-\widehat{R}(\w_\phi, \widehat{T}) \le 0.
    \end{align*}    
    The proof concludes by bounding the third term of  Eq. \eqref{eq:generalization_bound_decomposition} using Lemma \ref{lem:empirical_risk_estimated_T_vs_oracle_T},
    \begin{align*}
        | \widehat{R}_\phi(\w_\phi, \widehat{T}) - \widehat{R}_\phi(\w_\phi, T_o)|
        =
        B E(n_0,d).
    \end{align*}
\end{proof}

%
%
\section{Simulation Results}
\label{simulations}
In this section we present the simulation results corresponding the various setups described in Section \ref{sec:results}. 

\begin{table}[htp!]
\begin{center}
\begin{tabular}{c}
\hspace{-3in}
\begin{minipage}{0.25\columnwidth}
 \scalebox{0.63}{
 \begin{tabular}{c|c|c|r r r r r r r r} \hline \hline
                Marginals &  Class labels  &  $n$   & \multicolumn{1}{c}{SLB} & \multicolumn{1}{c}{LU} & \multicolumn{1}{c}{TAN} &  \multicolumn{1}{c}{BN} & \multicolumn{1}{c}{RF} & \multicolumn{1}{c}{SVM} & \multicolumn{1}{c}{5-NN} \\ \hline
                \multirow{8}{*}{Normal} &  \multirow{4}{*}{Balanced} &      200   &  $ 15.4 \pm 2.98 $ &  $ 21.2 \pm 3.64 $ & ${\bf 8.17 \pm 3.78 }$ & $ 23 \pm 3.75 $ & $ 15 \pm 2.54 $ & $ 11.9 \pm 2.36 $ & $ 13.8 \pm 2.65 $\\
		&&	400   &  $ 9.37 \pm 2.34 $ &  $ 19.4 \pm 3.5 $ & ${\bf 6.68 \pm 3.88 }$ & $ 21.8 \pm 3.66 $ & $ 11.6 \pm 2.08 $ & $ 8.07 \pm 2.02 $ & $ 11.4 \pm 2.56 $\\
		&&	600   &  $ 6.91 \pm 1.76 $ &  $ 17.6 \pm 3.45 $ & ${\bf 4.98 \pm 2.64 }$ & $ 20.3 \pm 3.36 $ & $ 9.73 \pm 1.55 $ & $ 6.34 \pm 1.42 $ & $ 9.91 \pm 2.1 $\\
		&&	800   &  ${\bf 5.82 \pm 1.82 }$ &  $ 18.3 \pm 3.19 $ & ${\bf 5.56 \pm 4.02 }$ & $ 21 \pm 3.14 $ & $ 8.87 \pm 1.67 $ & ${\bf 5.46 \pm 1.54 }$ & $ 8.89 \pm 2.22 $\\
		&&	1000   &  ${\bf 5.11 \pm 1.36 }$ &  $ 17.3 \pm 3.46 $ & ${\bf 4.66 \pm 2.87 }$ & $ 20.7 \pm 3.44 $ & $ 8.23 \pm 1.63 $ & ${\bf 4.84 \pm 1.16 }$ & $ 8.37 \pm 1.93 $\\
    		\cline{2-10}                
                & \multirow{4}{*}{Unbalanced}      &       200   &  $ 20.4 \pm 4.7 $ & $ 24.6 \pm 4.57 $ & ${\bf 11.4 \pm 5.51 }$ & $ 26.1 \pm 4.41 $ & $ 30.3 \pm 4.6 $ & $ 25.8 \pm 4.83 $ & $ 24.3 \pm 3.89 $\\
		&&	400   &  $ 13.2 \pm 3.71 $ &  $ 22.7 \pm 4.88 $ & ${\bf 7.37 \pm 4.32 }$ & $ 25.1 \pm 4.29 $ & $ 25.5 \pm 4.08 $ & $ 18.7 \pm 3.43 $ & $ 19.2 \pm 3.72 $\\
		&&	600   &  $ 9.48 \pm 2.98 $ &  $ 21.5 \pm 4.04 $ & ${\bf 6.52 \pm 3.96 }$ & $ 23.7 \pm 4.25 $ & $ 22 \pm 3.65 $ & $ 14.7 \pm 2.79 $ & $ 17.2 \pm 3.71 $\\
		&&	800   &  $ 7.43 \pm 2 $ &  $ 21.3 \pm 3.99 $ & ${\bf 6.13 \pm 4.05 }$ & $ 23.4 \pm 3.74 $ & $ 20.3 \pm 2.71 $ & $ 12.7 \pm 2.43 $ & $ 15.9 \pm 2.95 $\\
		&&	1000   &  ${\bf 7.04 \pm 2.12 }$ & $ 20.5 \pm 3.6 $ & ${\bf 6.56 \pm 4.54 }$ & $ 22.6 \pm 3.86 $ & $ 18.9 \pm 2.61 $ & $ 11.6 \pm 2.23 $ & $ 15 \pm 3.18 $\\
		\hline
                \multirow{8}{*}{Complex} &  \multirow{4}{*}{Balanced} &	200   &  $ 33.3 \pm 3.23 $ & $ 33.3 \pm 3.23 $ & $ 35.5 \pm 3.16 $ & $ 33.1 \pm 3.38 $ & ${\bf 30.8 \pm 2.89 }$ & $ 37.2 \pm 2.92 $ & $ 38.2 \pm 2.38 $\\
		&&	400   &   $ 31.1 \pm 3.71 $ & $ 30.7 \pm 3.97 $ & $ 32.8 \pm 3.85 $ & $ 30.8 \pm 3.66 $ & ${\bf 26.8 \pm 2.87 }$ & $ 33.9 \pm 3.18 $ & $ 36.1 \pm 2.95 $\\
		&&	600   &   $ 29.4 \pm 3.34 $ & $ 29.5 \pm 3.46 $ & $ 30.8 \pm 4.93 $ & $ 29.7 \pm 3.37 $ & ${\bf 25.5 \pm 2.49 }$ & $ 31.7 \pm 3.19 $ & $ 34.8 \pm 2.91 $\\
		&&	800   &  $ 27.7 \pm 3.13 $ & $ 28.1 \pm 3.32 $ & $ 28.4 \pm 4.83 $ & $ 27.9 \pm 3.34 $ & ${\bf 24.2 \pm 2.28 }$ & $ 29.3 \pm 2.62 $ & $ 33.5 \pm 2.64 $\\
		&&	1000 & $ 27.6 \pm 3.33 $ & $ 28.1 \pm 3.65 $ & $ 27.8 \pm 4.79 $ & $ 28.1 \pm 3.68 $ & ${\bf 23.8 \pm 2.41 }$ & $ 28.7 \pm 2.61 $ & $ 33.1 \pm 2.65 $\\
                  \cline{2-10}                
                & \multirow{4}{*}{Unbalanced}      &       200   &   ${\bf 37.1 \pm 3.72 }$ & ${\bf 37.1 \pm 3.86 }$ & ${\bf 38 \pm 3.41 }$ & $ 37.9 \pm 3.35 $ & $ 43.2 \pm 3.13 $ & $ 48.3 \pm 1.57 $ & $ 43.8 \pm 1.75 $\\
		&&	400   &  $ 34.9 \pm 3.71 $ & $ 35.3 \pm 3.87 $ & ${\bf 33.3 \pm 4.05 }$ & $ 35.6 \pm 3.6 $ & $ 39.7 \pm 3.12 $ & $ 45.9 \pm 2.69 $ & $ 41.3 \pm 2.48 $\\
		&&	600   &  ${\bf 32.7 \pm 3.66 }$ & ${\bf 33 \pm 3.72 }$ & ${\bf 31.2 \pm 4.65 }$ & $ 33.6 \pm 3.58 $ & $ 37.6 \pm 3.07 $ & $ 43.7 \pm 2.93 $ & $ 39.9 \pm 2.26 $\\
		&&	800   &  ${\bf 32.2 \pm 3.59 }$ & $ 33.2 \pm 3.75 $ & ${\bf 30.5 \pm 4.71 }$ & $ 33.5 \pm 3.65 $ & $ 36.4 \pm 3.13 $ & $ 42.5 \pm 2.99 $ & $ 39.3 \pm 2.06 $\\
		&&	1000   &  $ 31.6 \pm 3.68 $ & $ 32.9 \pm 3.93 $ & ${\bf 29 \pm 5.04 }$ & $ 33.1 \pm 3.7 $ & $ 35.7 \pm 3.32 $ & $ 41.4 \pm 3.03 $ & $ 38.7 \pm 2.68 $\\
               \hline \hline
        \end{tabular}}
      \end{minipage}
\end{tabular}
\end{center}
\caption{Misclassification error rates for the various classifiers, in the forest setting, averaged across 100 replicate datasets generated at each factor level combination. 
The classifier with the best performance, as well as classifiers that achieve comparable error rates are printed in boldface (see main text for details).}
\label{fig:forest_Simulation_results}
\end{table}

\begin{figure}
\vspace{-0.25in}
    \begin{tabular}{cc}
         &\hspace{-2.8in}  Structure: Forest, \hspace{0.05in} Marginal: Normal, \hspace{0.05in} Prior: Balanced \\\\
        \hspace{-0.25in} $n=200$ & \hspace{-0.05in} $n=400$ \\
      \includegraphics[trim=0in 0in 0.01in 0.1in,clip,width=0.42\linewidth]{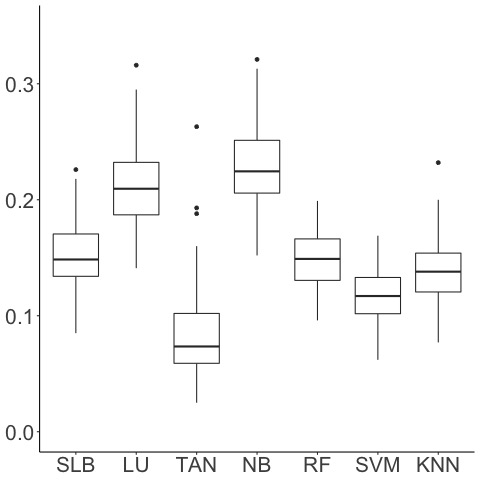}  \quad&  \quad
     \includegraphics[trim=0in 0in 0.01in 0.1in,clip,width=0.42\linewidth]{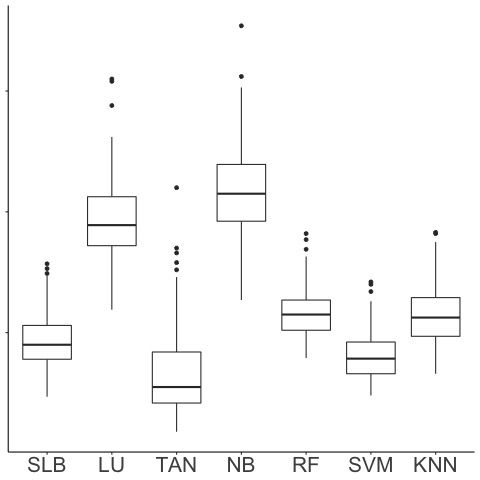} \\\\
     \hspace{-0.25in} $n=600$ & \hspace{-0.05in} $n=800$ \\
    \includegraphics[trim=0in 0in 0.01in 0.1in,clip,width=0.42\linewidth]{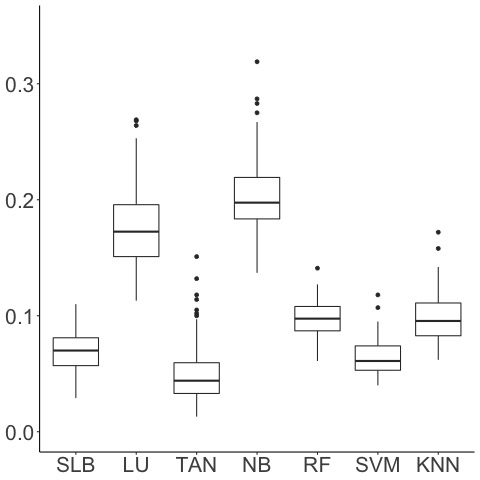} \quad &  \quad
     \includegraphics[trim=0in 0in 0.01in 0.1in,clip,width=0.42\linewidth]{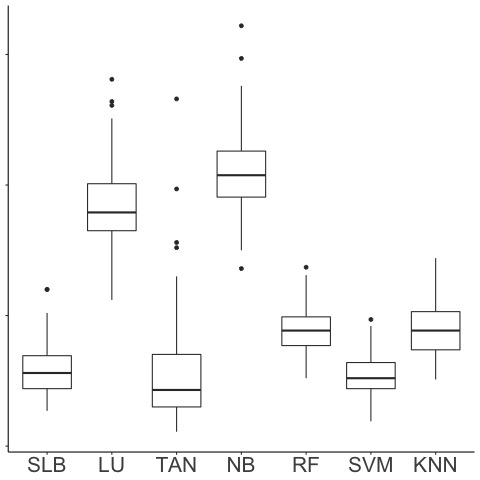}\\\\
     \hspace{-0.25in} $n=1000$ &  \\
    \includegraphics[trim=0in 0in 0.01in 0.1in,clip,width=0.42\linewidth]{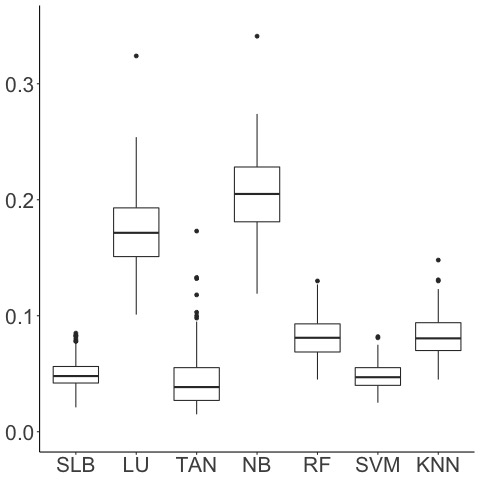} \quad &  \quad
    \begin{minipage}{0.45\columnwidth}
				\vspace{-3.7in}
     \caption{Misclassification error rates for the various classifiers, under the Gaussian forest regime and balanced training set, averaged across 100 replicate
                  datasets.}
      \label{forest_BNs_Gaussian_balanced}	
     \end{minipage}
          \end{tabular}
\end{figure}
\begin{figure}
\vspace{-0.25in}
    \begin{tabular}{cc}
         &\hspace{-2.8in}  Structure: Forest, \hspace{0.05in} Marginal: Normal, \hspace{0.05in} Prior: Imbalanced \\\\
        \hspace{-0.25in} $n=200$ & \hspace{-0.05in} $n=400$ \\
      \includegraphics[trim=0in 0in 0.01in 0.1in,clip,width=0.42\linewidth]{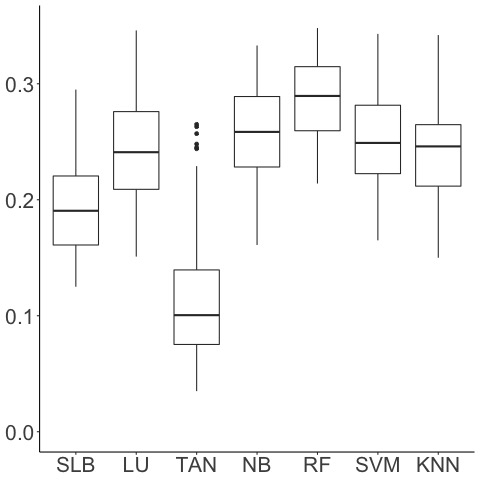}  \quad&  \quad
     \includegraphics[trim=0in 0in 0.01in 0.1in,clip,width=0.42\linewidth]{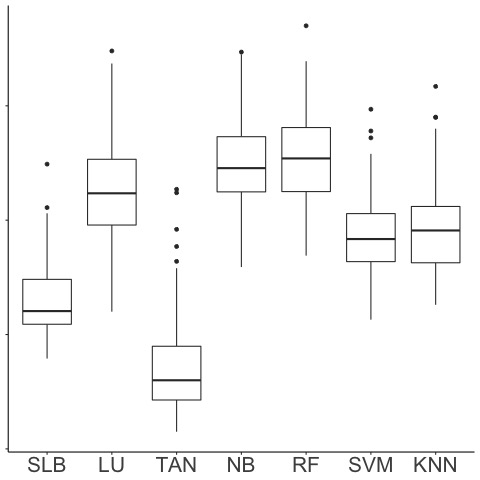} \\\\
     \hspace{-0.25in} $n=600$ & \hspace{-0.05in} $n=800$ \\
    \includegraphics[trim=0in 0in 0.01in 0.1in,clip,width=0.42\linewidth]{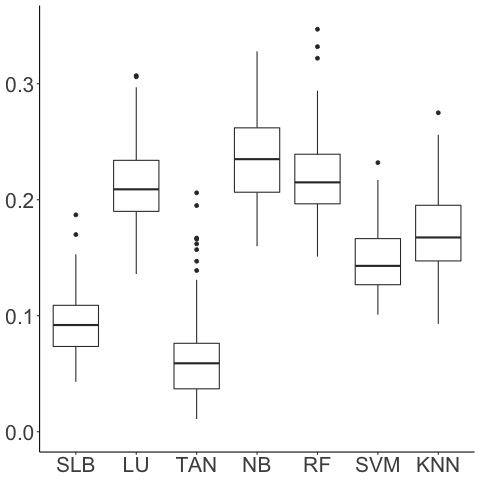} \quad &  \quad
     \includegraphics[trim=0in 0in 0.01in 0.1in,clip,width=0.42\linewidth]{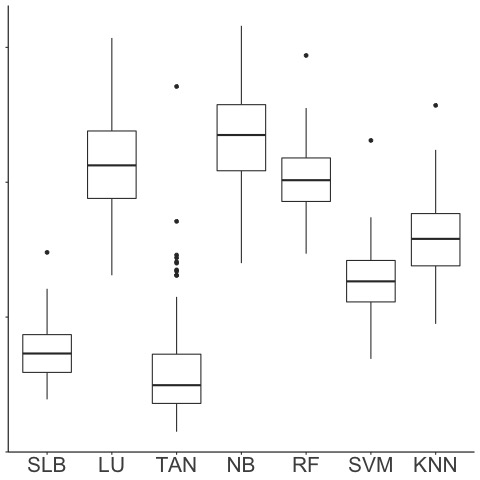}\\\\
     \hspace{-0.25in} $n=1000$ &  \\
    \includegraphics[trim=0in 0in 0.01in 0.1in,clip,width=0.42\linewidth]{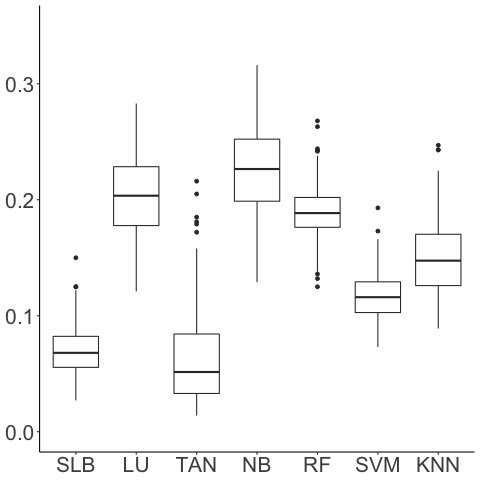} \quad &  \quad
    \begin{minipage}{0.45\columnwidth}
				\vspace{-3.7in}
     \caption{Misclassification error rates for the various classifiers, under the Gaussian forest regime and imbalanced training set, averaged across 100 replicate
                  datasets.}
				\label{forest_BNs_Gaussian_imbalanced}	
			\end{minipage}
          \end{tabular}
\end{figure}
\begin{figure}
\vspace{-0.25in}
    \begin{tabular}{cc}
         &\hspace{-2.8in}  Structure: Forest, \hspace{0.05in} Marginal: Complex, \hspace{0.05in} Prior: Balanced \\\\
        \hspace{-0.25in} $n=200$ & \hspace{-0.05in} $n=400$ \\
      \includegraphics[trim=0in 0in 0.01in 0.1in,clip,width=0.42\linewidth]{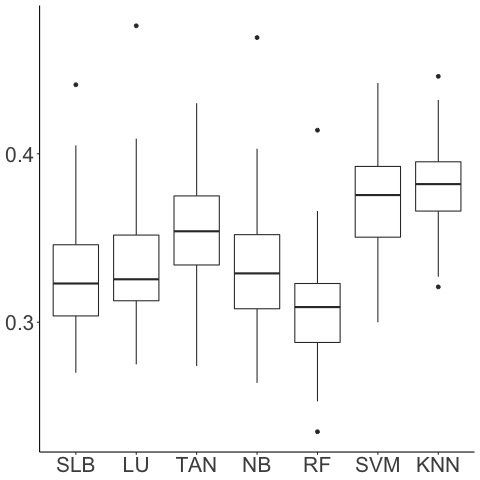}  \quad&  \quad
     \includegraphics[trim=0in 0in 0.01in 0.1in,clip,width=0.42\linewidth]{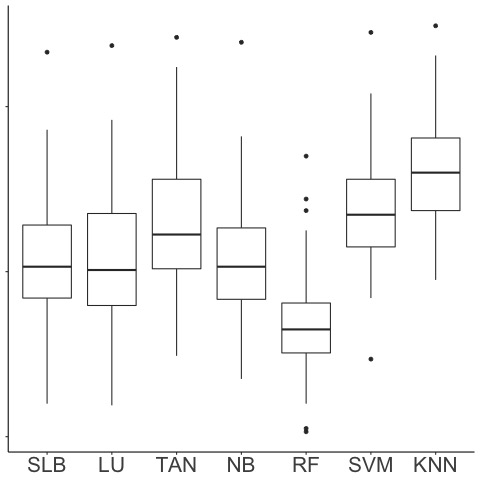} \\\\
     \hspace{-0.25in} $n=600$ & \hspace{-0.05in} $n=800$ \\
    \includegraphics[trim=0in 0in 0.01in 0.1in,clip,width=0.42\linewidth]{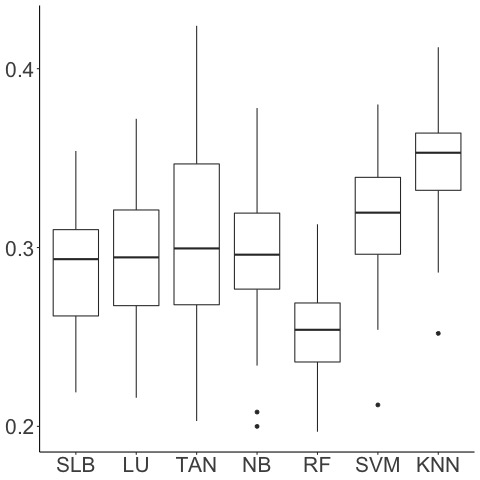} \quad &  \quad
     \includegraphics[trim=0in 0in 0.01in 0.1in,clip,width=0.42\linewidth]{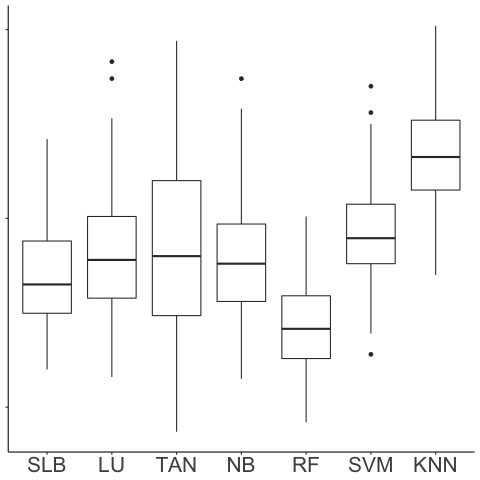}\\\\
     \hspace{-0.25in} $n=1000$ &  \\
    \includegraphics[trim=0in 0in 0.01in 0.1in,clip,width=0.42\linewidth]{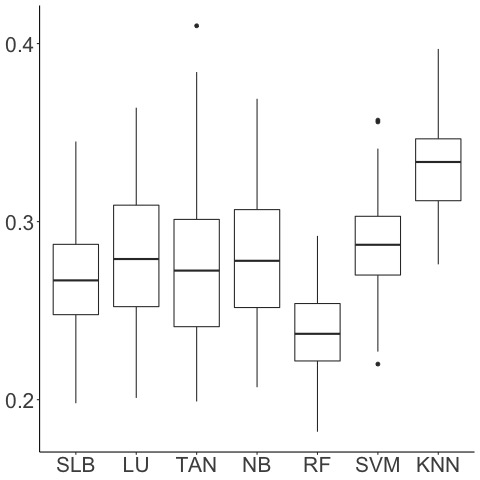} \quad &  \quad
    \begin{minipage}{0.45\columnwidth}
				\vspace{-3.7in}
     \caption{Misclassification error rates for the various classifiers, under the complex forest regime and balanced training set, averaged across 100 replicate
                  datasets.}
	\label{forest_BNs_complex_balanced}	
			\end{minipage}
          \end{tabular}
\end{figure}
\begin{figure}
\vspace{-0.25in}
    \begin{tabular}{cc}
         &\hspace{-2.8in}  Structure: Forest, \hspace{0.05in} Marginal: Complex, \hspace{0.05in} Prior: Imbalanced \\\\
        \hspace{-0.25in} $n=200$ & \hspace{-0.05in} $n=400$ \\
      \includegraphics[trim=0in 0in 0.01in 0.1in,clip,width=0.42\linewidth]{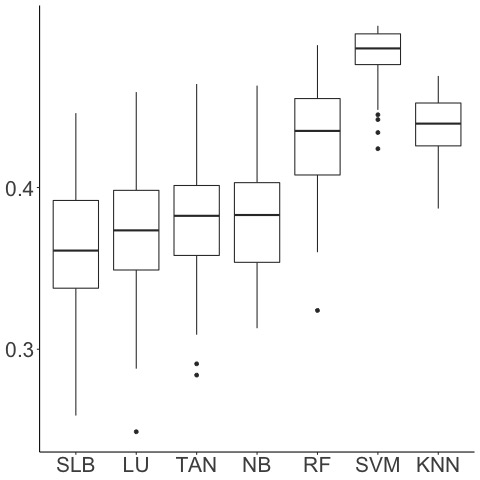}  \quad&  \quad
     \includegraphics[trim=0in 0in 0.01in 0.1in,clip,width=0.42\linewidth]{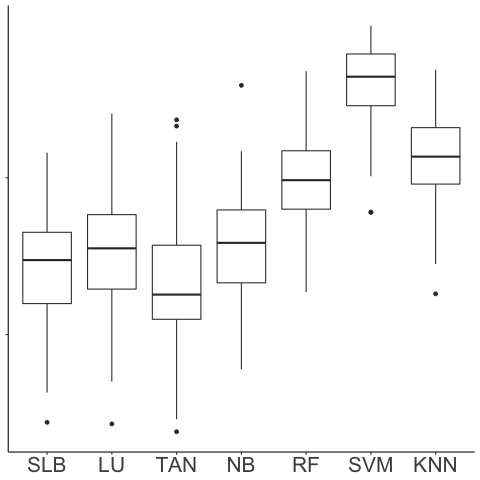} \\\\
     \hspace{-0.25in} $n=600$ & \hspace{-0.05in} $n=800$ \\
    \includegraphics[trim=0in 0in 0.01in 0.1in,clip,width=0.42\linewidth]{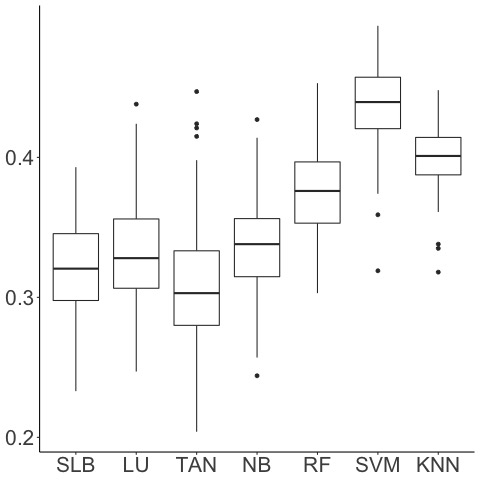} \quad &  \quad
     \includegraphics[trim=0in 0in 0.01in 0.1in,clip,width=0.42\linewidth]{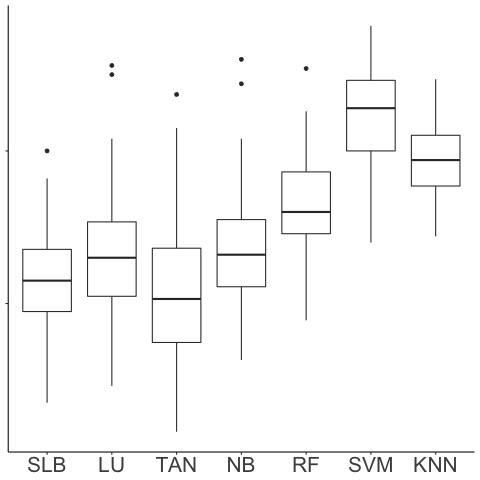}\\\\
     \hspace{-0.25in} $n=1000$ &  \\
    \includegraphics[trim=0in 0in 0.01in 0.1in,clip,width=0.42\linewidth]{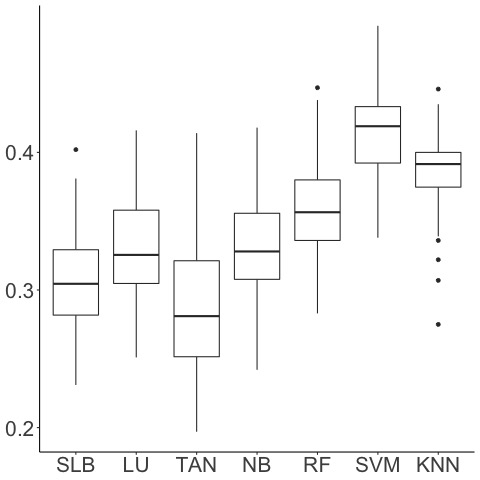} \quad &  \quad
    \begin{minipage}{0.45\columnwidth}
				\vspace{-3.7in}
     \caption{Misclassification error rates for the various classifiers, under the complex forest regime and imbalanced training set, averaged across 100 replicate
                  datasets.}
     \label{forest_BNs_complex_imbalanced}	
			\end{minipage}
          \end{tabular}
\end{figure}

\begin{table}[htp!]
\begin{center}
\begin{tabular}{c}
\hspace{-3.75in}
\begin{minipage}{0.25\columnwidth}
 \scalebox{0.7}{
 \setlength\tabcolsep{3.5pt}
 \begin{tabular}{c|c|c|c|rrrrrrrr} \hline \hline
                &Common&&&&&&&&&\\
                Marginals&structure &  Class labels  &  $n$   &  \multicolumn{1}{c}{SLB-} &\multicolumn{1}{c}{SLB} & \multicolumn{1}{c}{LU} & \multicolumn{1}{c}{TAN} &  \multicolumn{1}{c}{BN} & \multicolumn{1}{c}{RF} & \multicolumn{1}{c}{SVM} & \multicolumn{1}{c}{5-NN} \\ \hline
                \multirow{16}{*}{Normal} & \multirow{8}{*}{None} & \multirow{4}{*}{Balanced}
                 &	200   &  $ 12.3 \pm 3.23 $ & ${\bf 11.7 \pm 2.92 }$ & $ 15.2 \pm 4.33 $ & $ 16.8 \pm 6.81 $ & $ 28.7 \pm 6.45 $ & $ 15 \pm 3.68 $ & $ 17.8 \pm 2.66 $ & $ 17.1 \pm 3.09 $\\
		&&	&	400   &  $ 7.53 \pm 2.1 $ & ${\bf 7.3 \pm 2.04 }$ & $ 13.2 \pm 3.83 $ & $ 12 \pm 6.83 $ & $ 26.4 \pm 6.13 $ & $ 10.3 \pm 2.82 $ & $ 12.5 \pm 2.74 $ & $ 13 \pm 2.76 $\\
		&&	&	600   &  $ 5.4 \pm 1.53 $ & ${\bf 5.24 \pm 1.46 }$ & $ 12.9 \pm 3.54 $ & $ 12.9 \pm 7.59 $ & $ 26.7 \pm 5.72 $ & $ 8.92 \pm 2.17 $ & $ 10.4 \pm 2.46 $ & $ 10.9 \pm 2.3 $\\
		&&	&	800   &  ${\bf 4.24 \pm 1.31 }$ & ${\bf 4.21 \pm 1.32 }$ & $ 11.7 \pm 3.3 $ & $ 9.96 \pm 5.22 $ & $ 24.4 \pm 6.18 $ & $ 7.47 \pm 2.24 $ & $ 8.83 \pm 2.49 $ & $ 9.26 \pm 2.09 $\\
		&&	&	1000   &  ${\bf 3.73 \pm 1.22 }$ & ${\bf 3.67 \pm 1.24 }$ & $ 11.7 \pm 3.79 $ & $ 10.8 \pm 7.81 $ & $ 24.8 \pm 6.82 $ & $ 6.5 \pm 1.91 $ & $ 7.75 \pm 2.33 $ & $ 8.89 \pm 2.47 $\\
		\cline{3-12}                
                && \multirow{4}{*}{Unbalanced}      
                &	200   &  ${\bf 15.1 \pm 4.52 }$ & ${\bf 15.1 \pm 3.73 }$ & $ 17.1 \pm 4.27 $ & ${\bf 16.2 \pm 8.36 }$ & $ 29.6 \pm 6.7 $ & $ 25.7 \pm 6.07 $ & $ 30.1 \pm 5.86 $ & $ 26.1 \pm 5.24 $\\
		&&	&	400   &  $ 10.4 \pm 2.61 $ &  ${\bf 8.87 \pm 2.51 }$ &  $ 14.9 \pm 3.95 $ & $ 14.4 \pm 8.41 $ & $ 27.8 \pm 6.93 $ & $ 20.3 \pm 5.39 $ & $ 25.3 \pm 4.21 $ & $ 20.5 \pm 4.43 $\\
		&&	&	600   &  $ 7.69 \pm 2.12 $ &  ${\bf 6.31 \pm 1.72 }$ &  $ 13.3 \pm 3.94 $ & $ 12.7 \pm 7.66 $ & $ 25.8 \pm 5.67 $ & $ 16.9 \pm 4.44 $ & $ 20.9 \pm 3.81 $ & $ 17.6 \pm 4.01 $\\
		&&	&	800   &  $ 6.21 \pm 1.78 $ &  ${\bf 4.99 \pm 1.77 }$ &  $ 13.3 \pm 4.02 $ & $ 10.5 \pm 6.41 $ & $ 26.1 \pm 5.41 $ & $ 16 \pm 4.21 $ & $ 19 \pm 2.93 $ & $ 16.2 \pm 3.31 $\\
		&&	&	1000   &  $ 5.29 \pm 1.54 $ &.${\bf 4.11 \pm 1.18 }$ &  $ 13 \pm 4.09 $ & $ 10.7 \pm 6.8 $ & $ 25 \pm 5.01 $ & $ 13.9 \pm 3.87 $ & $ 18.2 \pm 3.45 $ & $ 14.4 \pm 3.44 $\\
                \cline{2-12}   
                &\multirow{8}{*}{$\frac{1}{3}$ Shared} & \multirow{4}{*}{Balanced}
                &	200   &  $ 14.8 \pm 3.71 $ & ${\bf 14.3 \pm 3.59 }$ & $ 17.9 \pm 4.57 $ & $ 18.7 \pm 8.01 $ & $ 31.2 \pm 6.77 $ & $ 17.6 \pm 4.31 $ & $ 20.3 \pm 3.71 $ & $ 21.4 \pm 4.26 $\\
		&&	&	400   &  ${\bf 9.56 \pm 2.59 }$ & ${\bf 9.42 \pm 2.53 }$ & $ 15.5 \pm 4.93 $ & $ 16.5 \pm 7.97 $ & $ 29 \pm 7.08 $ & $ 12.6 \pm 3.42 $ & $ 14.8 \pm 3.03 $ & $ 15.9 \pm 3.52 $\\
		&&	&	600   &  ${\bf 7.02 \pm 1.79 }$ & ${\bf 6.92 \pm 1.74 }$ & $ 14.6 \pm 4.24 $ & $ 13.8 \pm 7.56 $ & $ 27.7 \pm 6.55 $ & $ 10.5 \pm 2.62 $ & $ 11.7 \pm 3.05 $ & $ 13.7 \pm 3.13 $\\
		&&	&	800   &  ${\bf 5.91 \pm 1.73 }$ & ${\bf 5.82 \pm 1.69 }$ & $ 13.7 \pm 4.21 $ & $ 14.1 \pm 7.37 $ & $ 26.7 \pm 6.21 $ & $ 9.28 \pm 2.8 $ & $ 10.7 \pm 2.45 $ & $ 12.5 \pm 2.98 $\\
		&&	&	1000   &  ${\bf 5.59 \pm 1.85 }$ & ${\bf 5.57 \pm 1.9 }$ & $ 14.2 \pm 4.03 $ & $ 14.4 \pm 7.91 $ & $ 27.6 \pm 6.39 $ & $ 8.78 \pm 2.77 $ & $ 9.61 \pm 2.56 $ & $ 12.1 \pm 3.15 $\\
                \cline{3-12}           
                && \multirow{4}{*}{Unbalanced} 
                 &	200   &  $ 19 \pm 5.44 $ & ${\bf 17.7 \pm 5.18 }$ & $ 21.4 \pm 5.66 $ & ${\bf 21.2 \pm 9.18 }$ & $ 32.9 \pm 6.1 $ & $ 29.5 \pm 6.45 $ & $ 35.2 \pm 6.87 $ & $ 31 \pm 4.89 $\\
		&&	&	400   &  $ 12 \pm 3.52 $ & ${\bf 11.6 \pm 3.33 }$ & $ 18.8 \pm 5.04 $ & $ 19.6 \pm 9.23 $ & $ 31.2 \pm 6.95 $ & $ 24 \pm 5.6 $ & $ 28.3 \pm 5.26 $ & $ 25.6 \pm 4.79 $\\
		&&	&	600   &  $ 9.16 \pm 2.87 $ & ${\bf 8.97 \pm 2.82 }$ & $ 18.1 \pm 5.73 $ & $ 17.1 \pm 8.19 $ & $ 29.7 \pm 6.59 $ & $ 21.1 \pm 5.89 $ & $ 24.7 \pm 5.29 $ & $ 23 \pm 4.79 $\\
		&&	&	800   &  $ 7.96 \pm 2.24 $ & ${\bf 7.8 \pm 2.16 }$ & $ 18 \pm 5.4 $ & $ 17.1 \pm 8.3 $ & $ 30.2 \pm 5.85 $ & $ 20 \pm 5.04 $ & $ 22.9 \pm 4.03 $ & $ 21.3 \pm 3.95 $\\
		&&	&	1000   &  ${\bf 6.77 \pm 2.2 }$ & ${\bf 6.67 \pm 2.21 }$ & $ 17.7 \pm 5.45 $ & $ 16 \pm 8.13 $ & $ 29.8 \pm 6.65 $ & $ 18.4 \pm 5.49 $ & $ 21 \pm 4.28 $ & $ 20 \pm 4.73 $\\
                \hline
                \cline{3-12} \multirow{16}{*}{Complex} &  \multirow{8}{*}{None} & \multirow{4}{*}{Balanced}
                &	200   &  $ 14 \pm 3.29 $ & ${\bf 13.5 \pm 3.13 }$ & $ 17.9 \pm 4.71 $ & ${\bf 16.9 \pm 7.79 }$ & $ 30.1 \pm 6.54 $ & $ 16.6 \pm 3.78 $ & $ 18.7 \pm 3.33 $ & $ 18.9 \pm 3.47 $\\
		&&	&	400   &  $ 8.54 \pm 1.84 $ & ${\bf 8.32 \pm 1.86 }$ & $ 15.3 \pm 3.71 $ & $ 13.6 \pm 6.91 $ & $ 27.7 \pm 5.61 $ & $ 11.8 \pm 2.69 $ & $ 13.3 \pm 2.67 $ & $ 14.6 \pm 2.45 $\\
		&&	&	600   &  ${\bf 6.37 \pm 1.65 }$ & ${\bf 6.24 \pm 1.69 }$ & $ 15.1 \pm 3.93 $ & $ 13.2 \pm 7.49 $ & $ 27.1 \pm 5.46 $ & $ 10 \pm 2.58 $ & $ 10.4 \pm 2.57 $ & $ 12.2 \pm 2.73 $\\
		&&	&	800   &  ${\bf 5.17 \pm 1.44 }$ & ${\bf 5.13 \pm 1.42 }$ & $ 14.5 \pm 3.63 $ & $ 13.4 \pm 7.52 $ & $ 27 \pm 5.52 $ & $ 8.86 \pm 2.12 $ & $ 9.01 \pm 2.37 $ & $ 11.1 \pm 2.24 $\\
		&&	&	1000   &  ${\bf 4.55 \pm 1.15 }$ & ${\bf 4.48 \pm 1.13 }$ & $ 13.5 \pm 3.96 $ & $ 11.2 \pm 7.45 $ & $ 25.8 \pm 6.1 $ & $ 7.76 \pm 1.95 $ & $ 8.08 \pm 2.31 $ & $ 10.3 \pm 2.13 $\\
                  \cline{3-12}                
                && \multirow{4}{*}{Unbalanced}      
                &	200   &  $ 18.3 \pm 5 $ & ${\bf 17.1 \pm 4.98 }$ & $ 21 \pm 4.73 $ & ${\bf 18.4 \pm 7.8 }$ & $ 31.4 \pm 5.6 $ & $ 29.4 \pm 5.07 $ & $ 32.4 \pm 5.79 $ & $ 29.3 \pm 4.43 $\\
		&&	&	400   &  $ 10.9 \pm 2.76 $ & ${\bf 10.6 \pm 2.83 }$ & $ 18.7 \pm 4.88 $ & $ 14.8 \pm 7.45 $ & $ 29.5 \pm 5.15 $ & $ 23.6 \pm 5.19 $ & $ 25.8 \pm 4.17 $ & $ 23.4 \pm 3.91 $\\
		&&	&	600   &  ${\bf 8.18 \pm 2.29 }$ & ${\bf 8 \pm 2.34 }$ & $ 18.1 \pm 5.08 $ & $ 13 \pm 8.36 $ & $ 29 \pm 5.8 $ & $ 20.6 \pm 4.82 $ & $ 23.2 \pm 4.31 $ & $ 20.6 \pm 4.09 $\\
		&&	&	800   &  $ 7.04 \pm 1.8 $ & ${\bf 6.89 \pm 1.82 }$ & $ 18.3 \pm 4.5 $ & $ 15 \pm 8.81 $ & $ 28.8 \pm 5.6 $ & $ 19.8 \pm 4.2 $ & $ 20.9 \pm 3.38 $ & $ 19.5 \pm 3.5 $\\
		&&	&	1000   &  ${\bf 5.77 \pm 1.62 }$ & ${\bf 5.69 \pm 1.61 }$ & $ 17.4 \pm 4.74 $ & $ 13.1 \pm 7.98 $ & $ 27.7 \pm 5.63 $ & $ 17.8 \pm 4.18 $ & $ 18.9 \pm 2.84 $ & $ 18.1 \pm 3.81 $\\
                 \cline{2-12}       
                &\multirow{8}{*}{$\frac{1}{3}$ Shared} & \multirow{4}{*}{Balanced}
                &	200   &  $ 17.3 \pm 4.02 $ & ${\bf 16.5 \pm 3.86 }$ & $ 20.9 \pm 5.39 $ & $ 19.7 \pm 7.28 $ & $ 32.8 \pm 6.65 $ & $ 19.4 \pm 4.26 $ & $ 21.1 \pm 3.89 $ & $ 22.5 \pm 3.8 $\\
		&&	&	400   &  $ 11.5 \pm 2.94 $ & ${\bf 11.1 \pm 2.92 }$ & $ 19 \pm 4.63 $ & $ 18.3 \pm 9 $ & $ 30.9 \pm 5.29 $ & $ 14.6 \pm 3.48 $ & $ 15.6 \pm 3.53 $ & $ 18.3 \pm 3.73 $\\
		&&	&	600   &  ${\bf 8.64 \pm 2.6 }$ & ${\bf 8.58 \pm 2.57 }$ & $ 16.7 \pm 4.61 $ & $ 15.8 \pm 8.03 $ & $ 28.6 \pm 6.21 $ & $ 11.8 \pm 3.06 $ & $ 13 \pm 3.11 $ & $ 16.1 \pm 3.51 $\\
		&&	&	800   &  ${\bf 7.43 \pm 2.22 }$ & ${\bf 7.36 \pm 2.17 }$ & $ 17.1 \pm 4.77 $ & $ 15.3 \pm 7.53 $ & $ 29.6 \pm 6.41 $ & $ 11 \pm 2.96 $ & $ 11.2 \pm 2.79 $ & $ 14.7 \pm 3.04 $\\
		&&	&	1000   &  ${\bf 6.38 \pm 1.73 }$ & ${\bf 6.34 \pm 1.7 }$ & $ 15.9 \pm 4.18 $ & $ 14.9 \pm 8.07 $ & $ 28 \pm 5.81 $ & $ 9.83 \pm 2.42 $ & $ 10.1 \pm 2.57 $ & $ 13.5 \pm 3.02 $\\
		 \cline{3-12}           
                && \multirow{4}{*}{Unbalanced} 
               &	200   &  $ 20.8 \pm 5.64 $ & ${\bf 19.7 \pm 5.58 }$ & $ 24.1 \pm 6.42 $ & ${\bf 23.3 \pm 8.8 }$ & $ 35.1 \pm 5.91 $ & $ 32.6 \pm 6.77 $ & $ 37.1 \pm 7.2 $ & $ 32.9 \pm 5 $\\
		&&	&	400   &  $ 14.2 \pm 3.76 $ & ${\bf 13.7 \pm 3.59 }$ & $ 21.6 \pm 5.17 $ & $ 19.9 \pm 9.47 $ & $ 32.5 \pm 6 $ & $ 26.8 \pm 5.86 $ & $ 29.7 \pm 5.18 $ & $ 27.4 \pm 4.23 $\\
		&&	&	600   &  $ 11.5 \pm 2.95 $ & ${\bf 11.2 \pm 2.78 }$ & $ 21.6 \pm 5.24 $ & $ 19.6 \pm 8.94 $ & $ 32.5 \pm 5.23 $ & $ 24.4 \pm 4.92 $ & $ 26.5 \pm 4.72 $ & $ 25.5 \pm 4.09 $\\
		&&	&	800   &  ${\bf 9.22 \pm 2.59 }$ & ${\bf 9.18 \pm 2.63 }$ & $ 20.5 \pm 5.99 $ & $ 17.8 \pm 9.27 $ & $ 30.8 \pm 6.51 $ & $ 22 \pm 5.65 $ & $ 23.6 \pm 4.14 $ & $ 23.1 \pm 4.47 $\\
		&&	&	1000   &  ${\bf 8.26 \pm 2.73 }$ & ${\bf 8.18 \pm 2.69 }$ & $ 20 \pm 5.79 $ & $ 17.2 \pm 8.01 $ & $ 30.8 \pm 5.53 $ & $ 20.6 \pm 5.7 $ & $ 21.7 \pm 4.25 $ & $ 21.9 \pm 4.49 $\\
                \hline \hline
        \end{tabular}}
      \end{minipage}
\end{tabular}
\end{center}
\caption{Misclassification error rates for the various classifiers, in the general BN setting, averaged across 100 replicate datasets generated at each factor level combination. The classifier with the best performance, as well as classifiers that achieve comparable error rates are printed in boldface (see main text for details).}
\label{fig:general_BNs_Simulation}
\end{table}

\begin{figure}
\vspace{-0.25in}
    \begin{tabular}{cc}
        &\hspace{-2.8in}  Structure: BN, \hspace{0.05in} Marginal: Normal, \hspace{0.05in} Prior: Balanced,  \hspace{0.05in} Common structure: No \\\\
        \hspace{-0.25in} $n=200$ & \hspace{-0.05in} $n=400$ \\
      \includegraphics[trim=0in 0in 0.01in 0.1in,clip,width=0.42\linewidth]{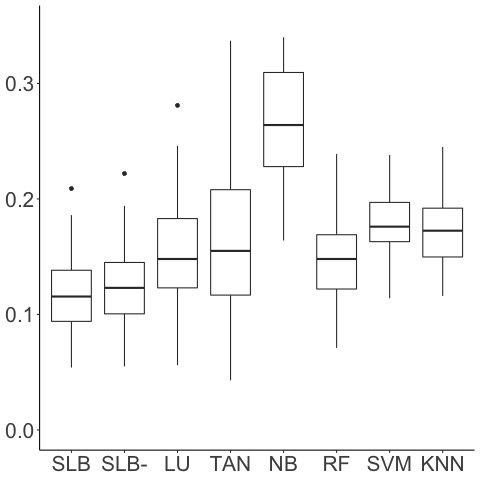}  \quad&  \quad
     \includegraphics[trim=0in 0in 0.01in 0.1in,clip,width=0.42\linewidth]{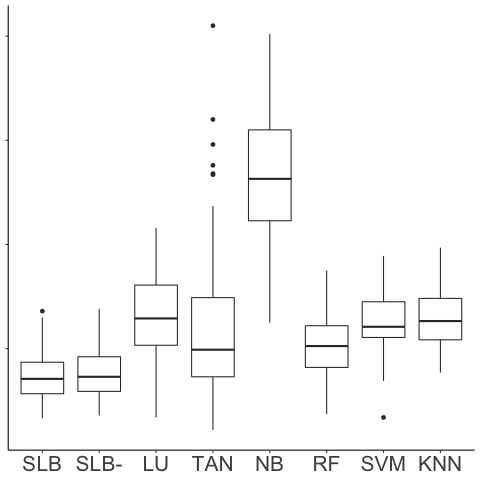} \\\\
     \hspace{-0.25in} $n=600$ & \hspace{-0.05in} $n=800$ \\
    \includegraphics[trim=0in 0in 0.01in 0.1in,clip,width=0.42\linewidth]{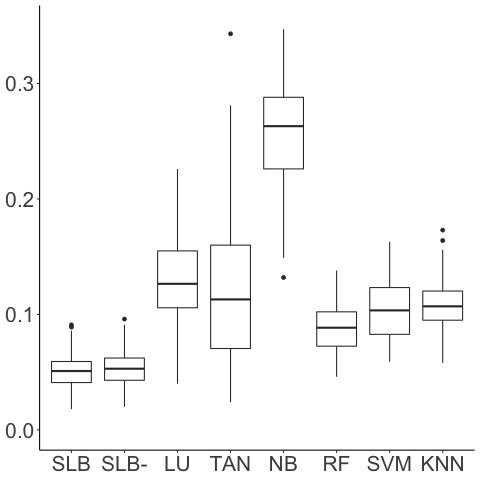} \quad &  \quad
     \includegraphics[trim=0in 0in 0.01in 0.1in,clip,width=0.42\linewidth]{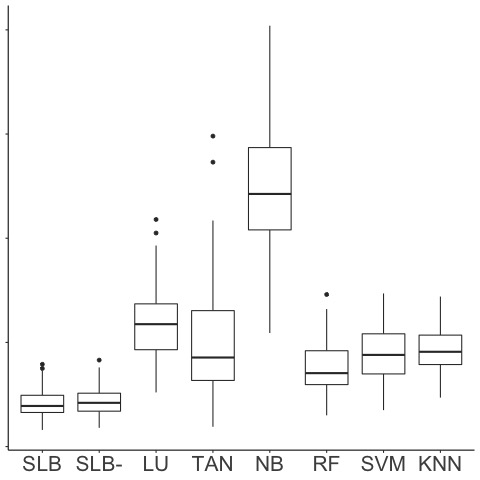}\\\\
     \hspace{-0.25in} $n=1000$ &  \\
    \includegraphics[trim=0in 0in 0.01in 0.1in,clip,width=0.42\linewidth]{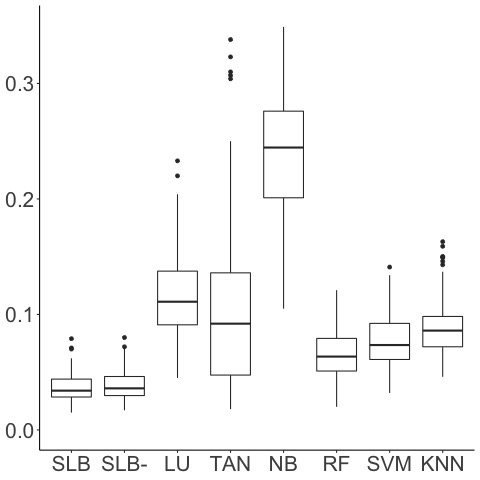} \quad &  \quad \hspace{0.1in}
    \begin{minipage}{0.5\columnwidth}
				\vspace{-3.7in}
     \caption{Misclassification error rates for the various classifiers, across 100 replicate datasets. As indicated, datasets are balanced  and sampled from Gaussian BNs with no common structure.}
       \label{general_BN_Gaussian}	             
		\end{minipage}
          \end{tabular}         
\end{figure}
\begin{figure}
\vspace{-0.25in}
    \begin{tabular}{cc}
        &\hspace{-2.8in}  Structure: BN, \hspace{0.05in} Marginal: Normal, \hspace{0.05in} Prior: Imbalanced,  \hspace{0.05in} Common structure: No \\\\
        \hspace{-0.25in} $n=200$ & \hspace{-0.05in} $n=400$ \\
      \includegraphics[trim=0in 0in 0.01in 0.1in,clip,width=0.42\linewidth]{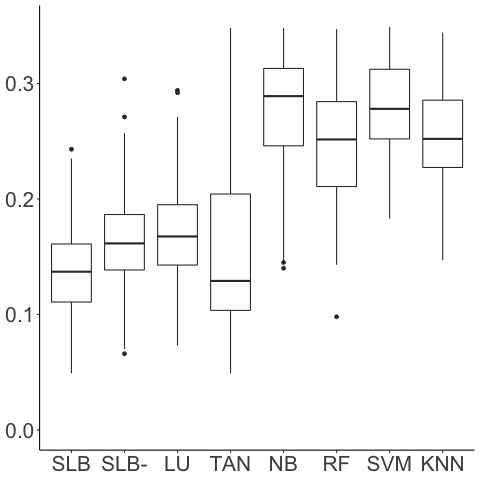}  \quad&  \quad
     \includegraphics[trim=0in 0in 0.01in 0.1in,clip,width=0.42\linewidth]{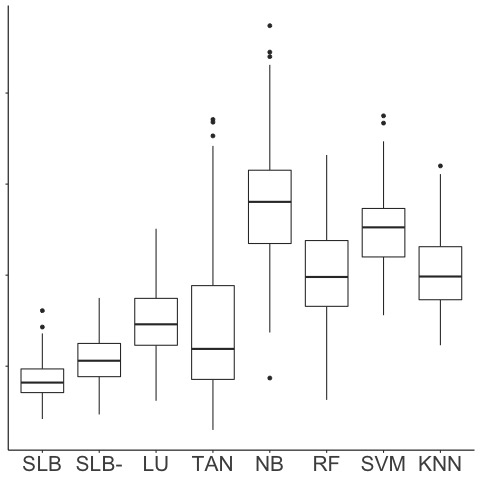} \\\\
     \hspace{-0.25in} $n=600$ & \hspace{-0.05in} $n=800$ \\
    \includegraphics[trim=0in 0in 0.01in 0.1in,clip,width=0.42\linewidth]{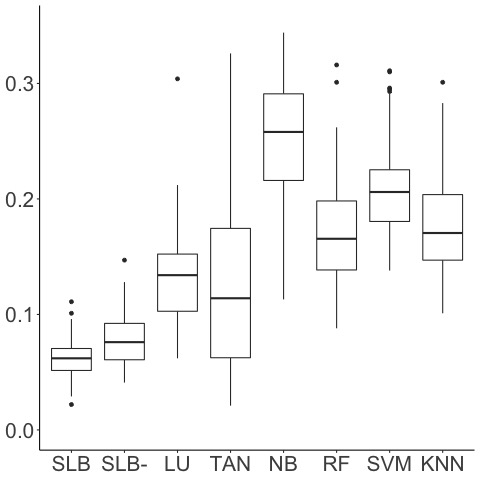} \quad &  \quad
     \includegraphics[trim=0in 0in 0.01in 0.1in,clip,width=0.42\linewidth]{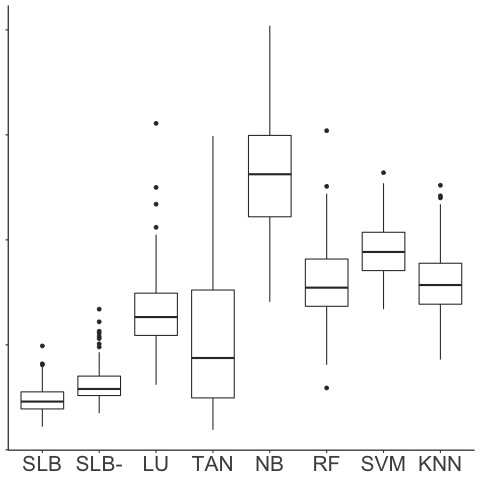}\\\\
     \hspace{-0.25in} $n=1000$ &  \\
    \includegraphics[trim=0in 0in 0.01in 0.1in,clip,width=0.42\linewidth]{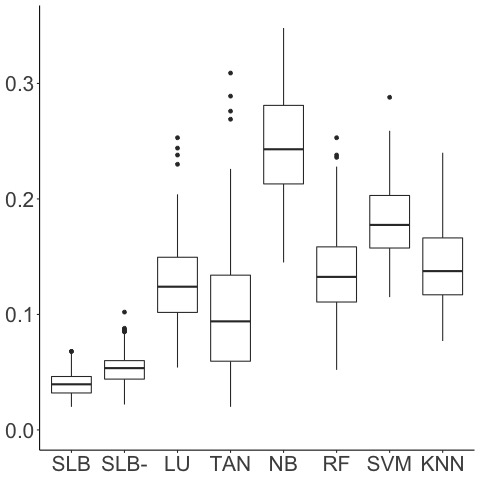} \quad &  \quad \hspace{0.1in}
    \begin{minipage}{0.5\columnwidth}
				\vspace{-3.7in}
     \caption{Misclassification error rates for the various classifiers, across 100 replicate datasets. As indicated, datasets are imbalanced  and sampled from Gaussian BNs with no common structure.}
				\label{example}	
			\end{minipage}
          \end{tabular}
\end{figure}
\begin{figure}
\vspace{-0.25in}
    \begin{tabular}{cc}
        &\hspace{-2.8in}  Structure: BN, \hspace{0.05in} Marginal: Normal, \hspace{0.05in} Prior: Balanced,  \hspace{0.05in} Common structure: Yes \\\\
        \hspace{-0.25in} $n=200$ & \hspace{-0.05in} $n=400$ \\
      \includegraphics[trim=0in 0in 0.01in 0.1in,clip,width=0.42\linewidth]{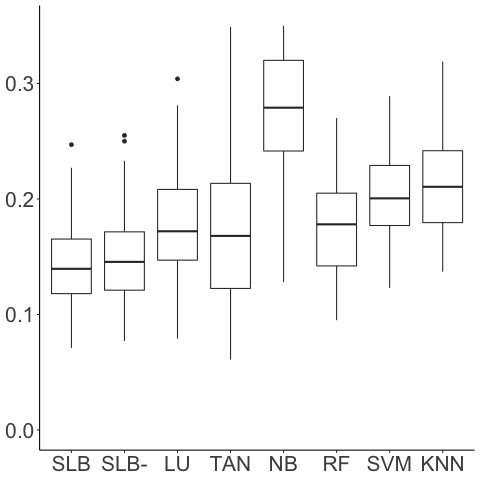}  \quad&  \quad
     \includegraphics[trim=0in 0in 0.01in 0.1in,clip,width=0.42\linewidth]{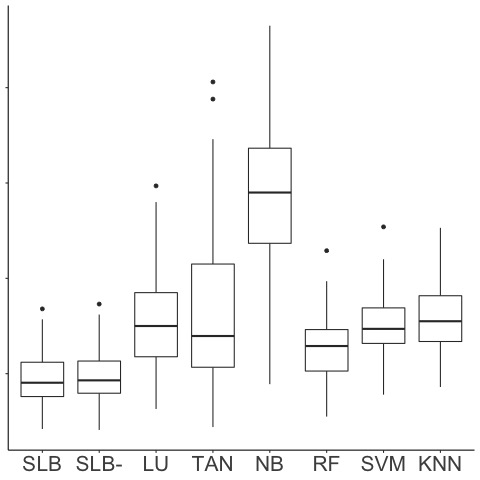} \\\\
     \hspace{-0.25in} $n=600$ & \hspace{-0.05in} $n=800$ \\
    \includegraphics[trim=0in 0in 0.01in 0.1in,clip,width=0.42\linewidth]{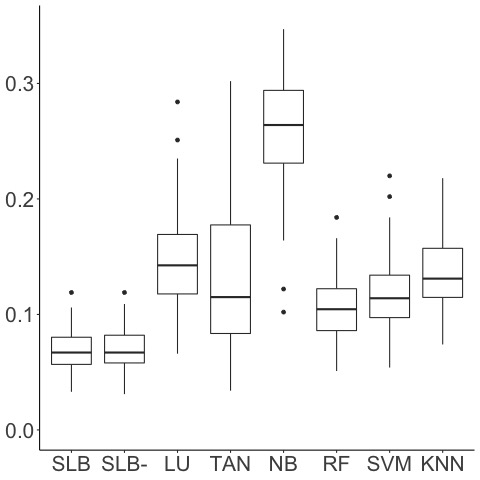} \quad &  \quad
     \includegraphics[trim=0in 0in 0.01in 0.1in,clip,width=0.42\linewidth]{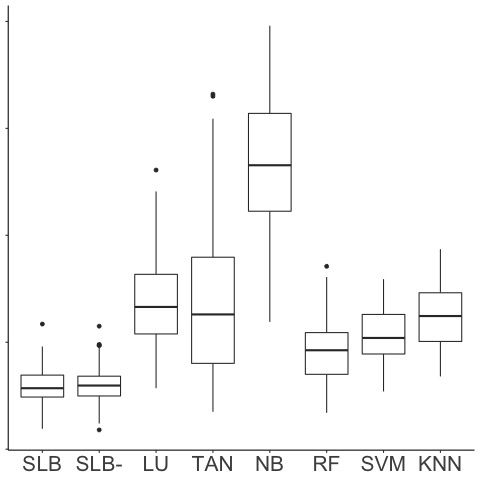}\\\\
     \hspace{-0.25in} $n=1000$ &  \\
    \includegraphics[trim=0in 0in 0.01in 0.1in,clip,width=0.42\linewidth]{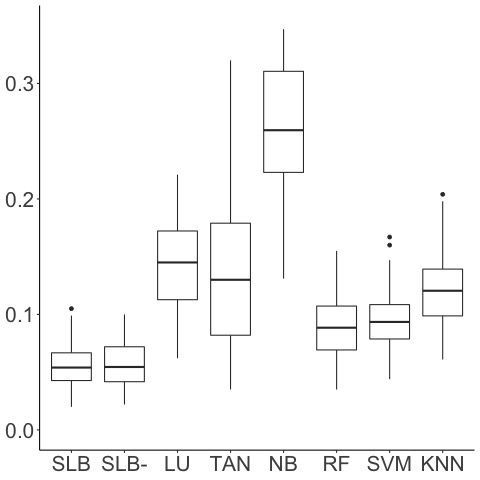} \quad &  \quad \hspace{0.1in}
    \begin{minipage}{0.5\columnwidth}				\
    \vspace{-3.7in}
     \caption{Misclassification error rates for the various classifiers, across 100 replicate datasets. As indicated, datasets are balanced  and sampled from Gaussian BNs with common structure.}
				\label{example}	
			\end{minipage}
          \end{tabular}
\end{figure}
\begin{figure}
\vspace{-0.25in}
    \begin{tabular}{cc}
        &\hspace{-2.8in}  Structure: BN, \hspace{0.05in} Marginal: Normal, \hspace{0.05in} Prior: Imbalanced,  \hspace{0.05in} Common structure: Yes \\\\
        \hspace{-0.25in} $n=200$ & \hspace{-0.05in} $n=400$ \\
      \includegraphics[trim=0in 0in 0.01in 0.1in,clip,width=0.42\linewidth]{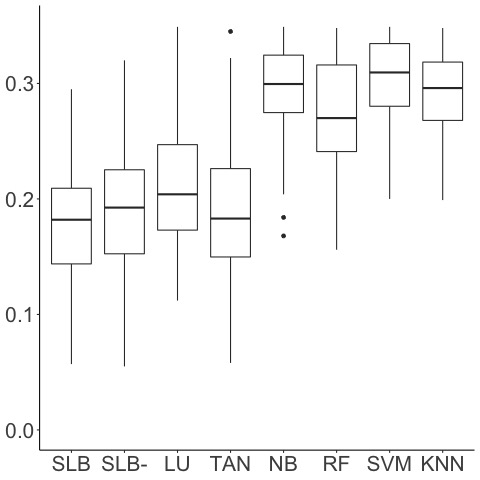}  \quad&  \quad
     \includegraphics[trim=0in 0in 0.01in 0.1in,clip,width=0.42\linewidth]{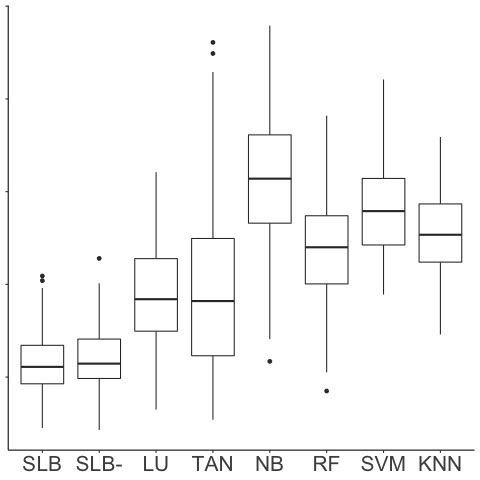} \\\\
     \hspace{-0.25in} $n=600$ & \hspace{-0.05in} $n=800$ \\
    \includegraphics[trim=0in 0in 0.01in 0.1in,clip,width=0.42\linewidth]{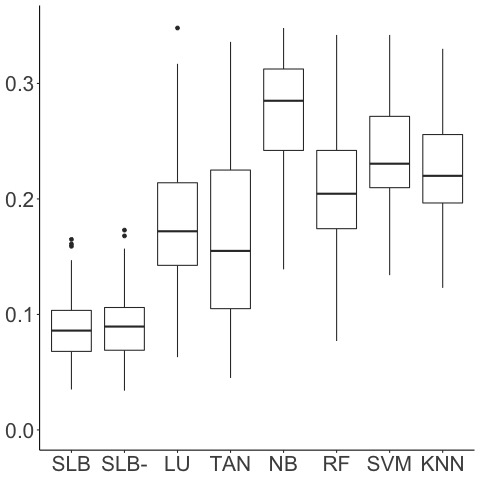} \quad &  \quad
     \includegraphics[trim=0in 0in 0.01in 0.1in,clip,width=0.42\linewidth]{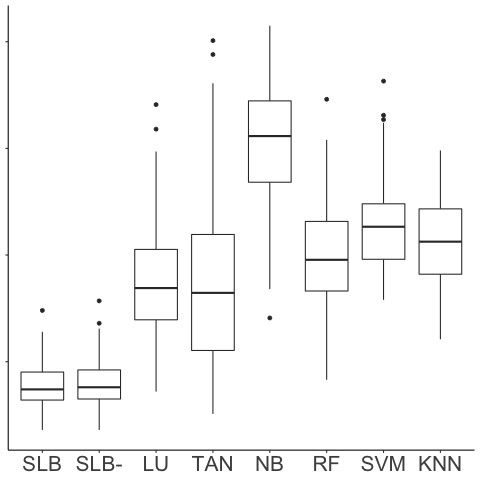}\\\\
     \hspace{-0.25in} $n=1000$ &  \\
    \includegraphics[trim=0in 0in 0.01in 0.1in,clip,width=0.42\linewidth]{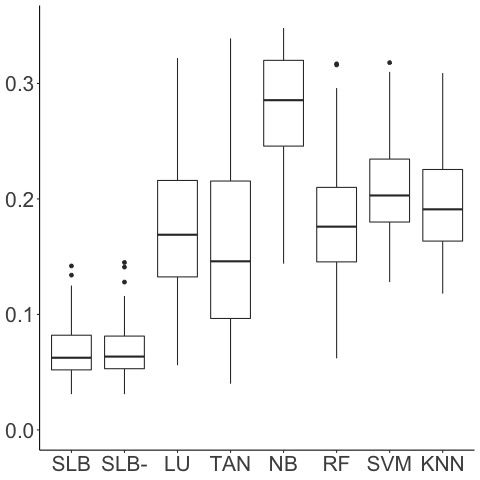} \quad &  \quad \hspace{0.1in}
    \begin{minipage}{0.5\columnwidth}
				\vspace{-3.7in}
     \caption{Misclassification error rates for the various classifiers, across 100 replicate datasets. As indicated, datasets are imbalanced  and sampled from Gaussian BNs with common structure.}
				\label{example}	
			\end{minipage}
          \end{tabular}
\end{figure}
\begin{figure}
\vspace{-0.25in}
    \begin{tabular}{cc}
        &\hspace{-2.8in}  Structure: BN, \hspace{0.05in} Marginal: Complex, \hspace{0.05in} Prior: Balanced,  \hspace{0.05in} Common structure: No \\\\
        \hspace{-0.25in} $n=200$ & \hspace{-0.05in} $n=400$ \\
      \includegraphics[trim=0in 0in 0.01in 0.1in,clip,width=0.42\linewidth]{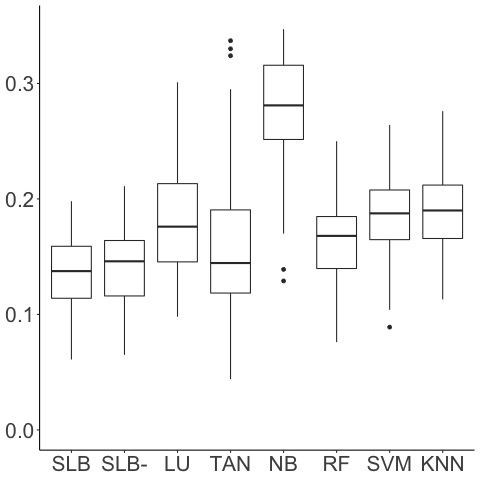}  \quad&  \quad
     \includegraphics[trim=0in 0in 0.01in 0.1in,clip,width=0.42\linewidth]{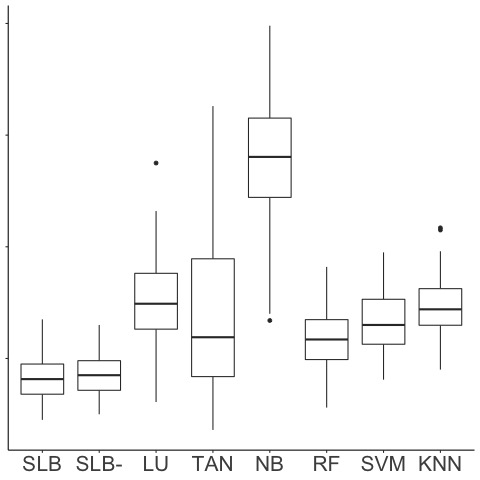} \\\\
     \hspace{-0.25in} $n=600$ & \hspace{-0.05in} $n=800$ \\
    \includegraphics[trim=0in 0in 0.01in 0.1in,clip,width=0.42\linewidth]{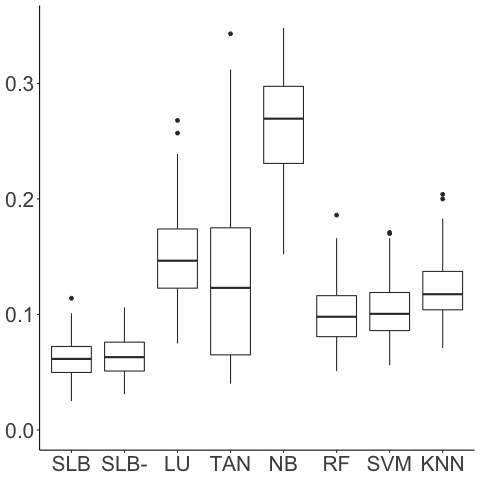} \quad &  \quad
     \includegraphics[trim=0in 0in 0.01in 0.1in,clip,width=0.42\linewidth]{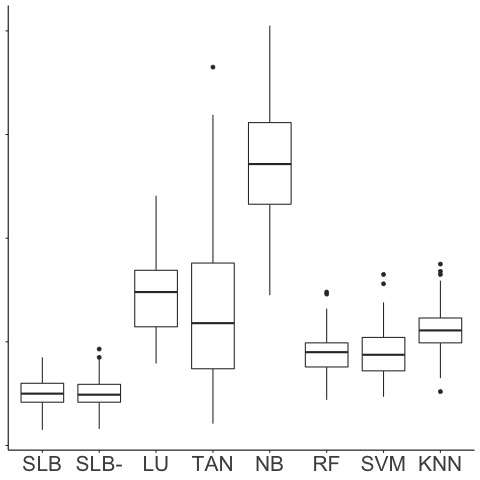}\\\\
     \hspace{-0.25in} $n=1000$ &  \\
    \includegraphics[trim=0in 0in 0.01in 0.1in,clip,width=0.42\linewidth]{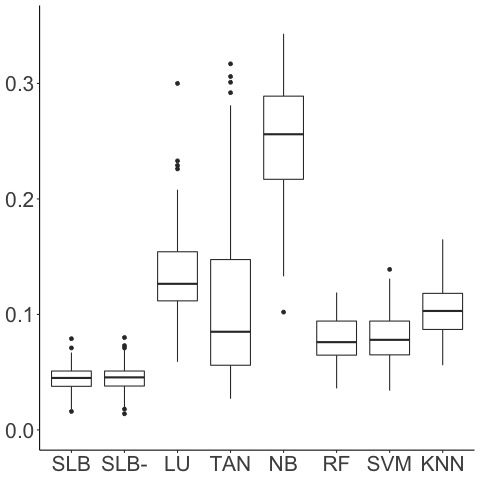} \quad &  \quad \hspace{0.1in}
    \begin{minipage}{0.5\columnwidth}
				\vspace{-3.7in}
				
     \caption{Misclassification error rates for the various classifiers, across 100 replicate datasets. As indicated, datasets are balanced and sampled from complex BNs with no common structure.}
				\label{example}	
			\end{minipage}
          \end{tabular}
\end{figure}
\begin{figure}
\vspace{-0.25in}
    \begin{tabular}{cc}
        &\hspace{-2.8in}  Structure: BN, \hspace{0.05in} Marginal: Complex, \hspace{0.05in} Prior: Imbalanced,  \hspace{0.05in} Common structure: No \\\\
        \hspace{-0.25in} $n=200$ & \hspace{-0.05in} $n=400$ \\
      \includegraphics[trim=0in 0in 0.01in 0.1in,clip,width=0.42\linewidth]{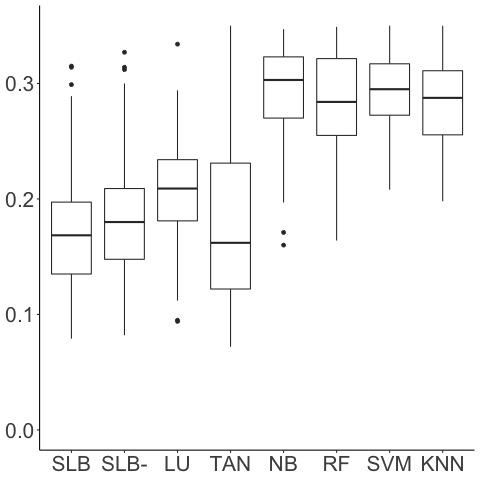}  \quad&  \quad
     \includegraphics[trim=0in 0in 0.01in 0.1in,clip,width=0.42\linewidth]{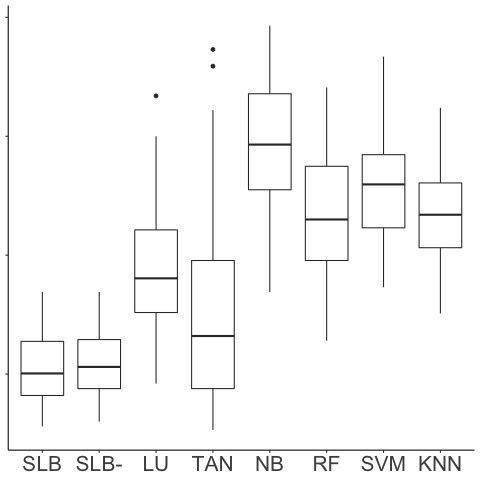} \\\\
     \hspace{-0.25in} $n=600$ & \hspace{-0.05in} $n=800$ \\
    \includegraphics[trim=0in 0in 0.01in 0.1in,clip,width=0.42\linewidth]{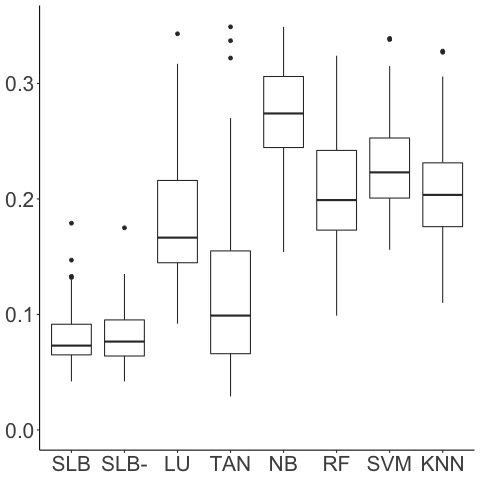} \quad &  \quad
     \includegraphics[trim=0in 0in 0.01in 0.1in,clip,width=0.42\linewidth]{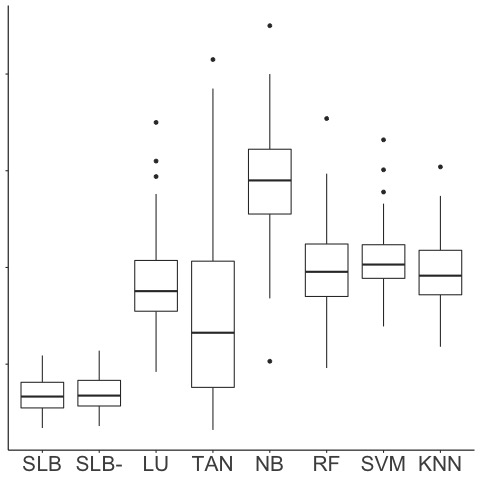}\\\\
     \hspace{-0.25in} $n=1000$ &  \\
    \includegraphics[trim=0in 0in 0.01in 0.1in,clip,width=0.42\linewidth]{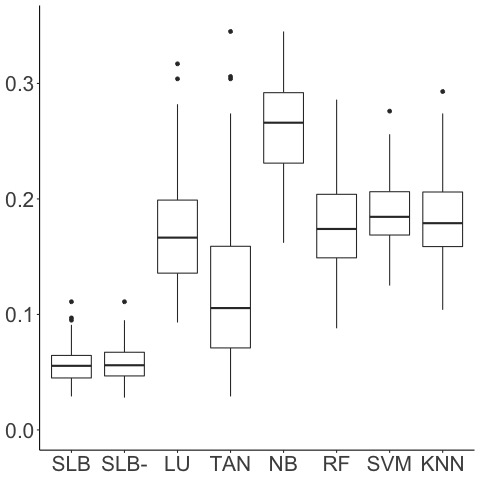} \quad &  \quad \hspace{0.1in}
    \begin{minipage}{0.5\columnwidth}
				\vspace{-3.7in}
     \caption{Misclassification error rates for the various classifiers, across 100 replicate datasets. As indicated, datasets are imbalanced and sampled from complex BNs with no common structure.}
				\label{example}	
			\end{minipage}
          \end{tabular}
\end{figure}
\begin{figure}
\vspace{-0.25in}
    \begin{tabular}{cc}
        &\hspace{-2.8in}  Structure: BN, \hspace{0.05in} Marginal: Complex, \hspace{0.05in} Prior: Balanced,  \hspace{0.05in} Common structure: Yes \\\\
        \hspace{-0.25in} $n=200$ & \hspace{-0.05in} $n=400$ \\
      \includegraphics[trim=0in 0in 0.01in 0.1in,clip,width=0.42\linewidth]{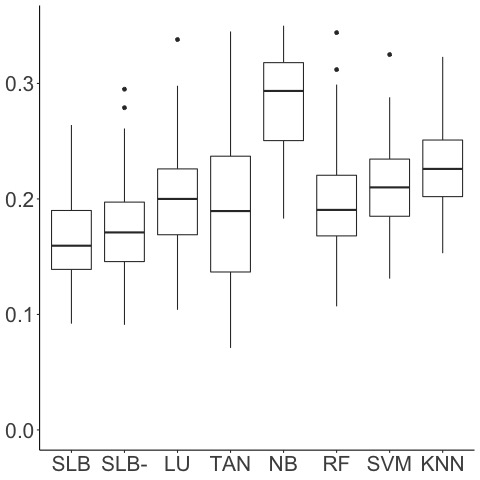}  \quad&  \quad
     \includegraphics[trim=0in 0in 0.01in 0.1in,clip,width=0.42\linewidth]{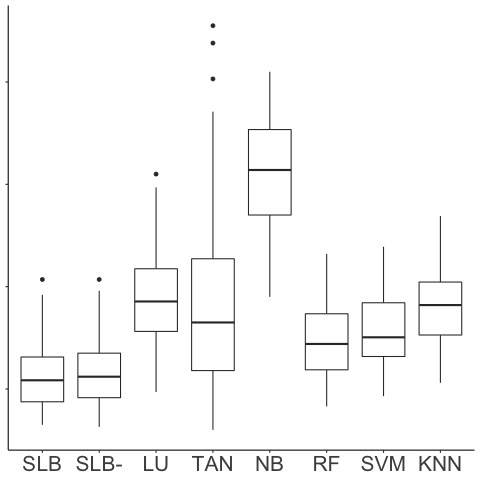} \\\\
     \hspace{-0.25in} $n=600$ & \hspace{-0.05in} $n=800$ \\
    \includegraphics[trim=0in 0in 0.01in 0.1in,clip,width=0.42\linewidth]{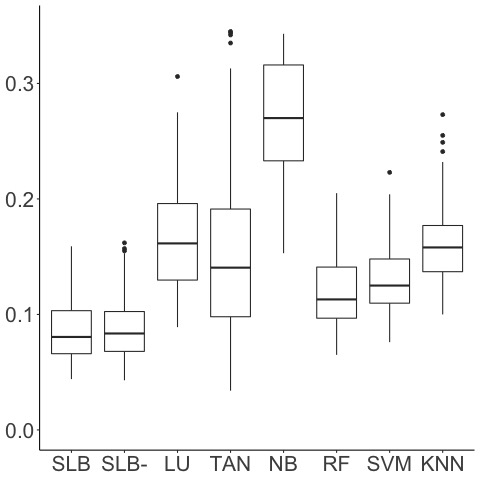} \quad &  \quad
     \includegraphics[trim=0in 0in 0.01in 0.1in,clip,width=0.42\linewidth]{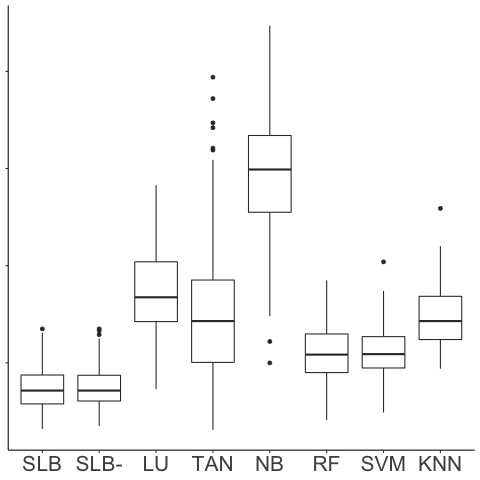}\\\\
     \hspace{-0.25in} $n=1000$ &  \\
    \includegraphics[trim=0in 0in 0.01in 0.1in,clip,width=0.42\linewidth]{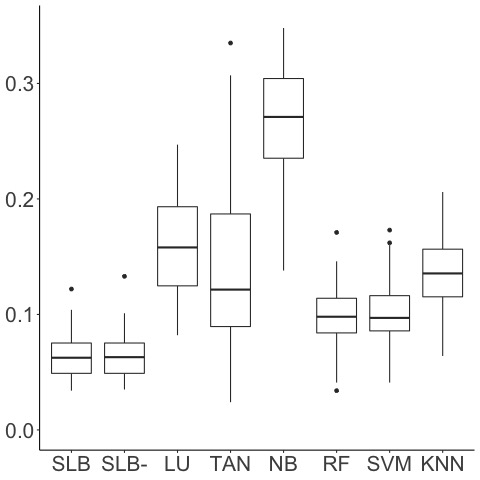} \quad &  \quad \hspace{0.1in}
    \begin{minipage}{0.5\columnwidth}
				\vspace{-3.7in}
				\centering
     \caption{Misclassification error rates for the various classifiers, across 100 replicate datasets. As indicated, datasets are balanced and sampled from complex BNs with common structure.}
				\label{example}	
			\end{minipage}
          \end{tabular}
\end{figure}
\begin{figure}
\vspace{-0.25in}
    \begin{tabular}{cc}
        &\hspace{-2.8in}  Structure: BN, \hspace{0.05in} Marginal: Complex, \hspace{0.05in} Prior: Imbalanced,  \hspace{0.05in} Common structure: Yes \\\\
        \hspace{-0.25in} $n=200$ & \hspace{-0.05in} $n=400$ \\
      \includegraphics[trim=0in 0in 0.01in 0.1in,clip,width=0.42\linewidth]{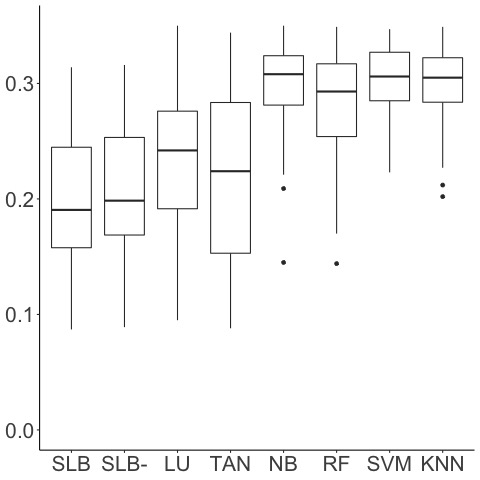}  \quad&  \quad
     \includegraphics[trim=0in 0in 0.01in 0.1in,clip,width=0.42\linewidth]{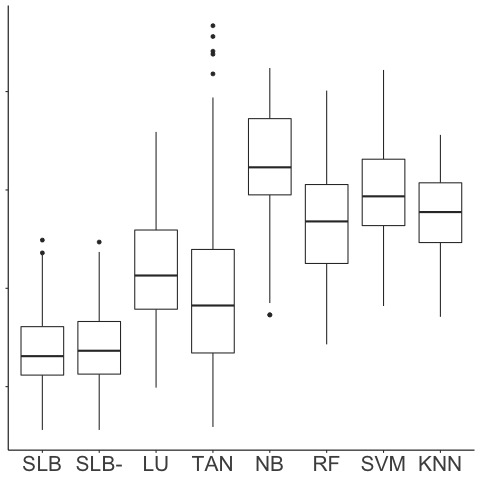} \\\\
     \hspace{-0.25in} $n=600$ & \hspace{-0.05in} $n=800$ \\
    \includegraphics[trim=0in 0in 0.01in 0.1in,clip,width=0.42\linewidth]{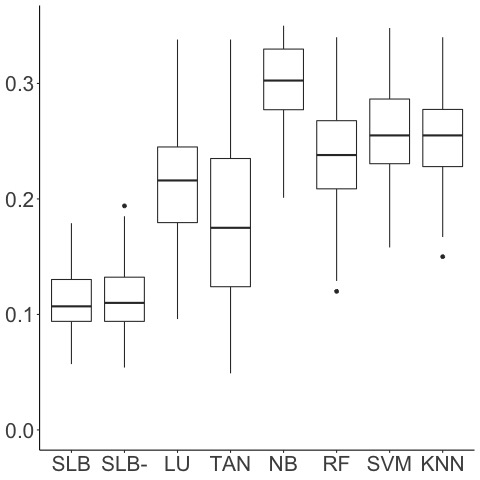} \quad &  \quad
     \includegraphics[trim=0in 0in 0.01in 0.1in,clip,width=0.42\linewidth]{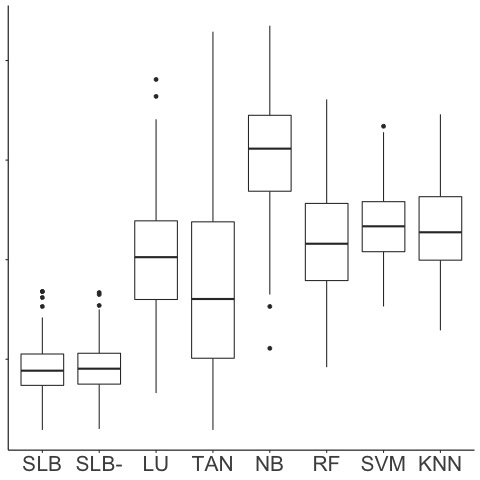}\\\\
     \hspace{-0.25in} $n=1000$ &  \\
    \includegraphics[trim=0in 0in 0.01in 0.1in,clip,width=0.42\linewidth]{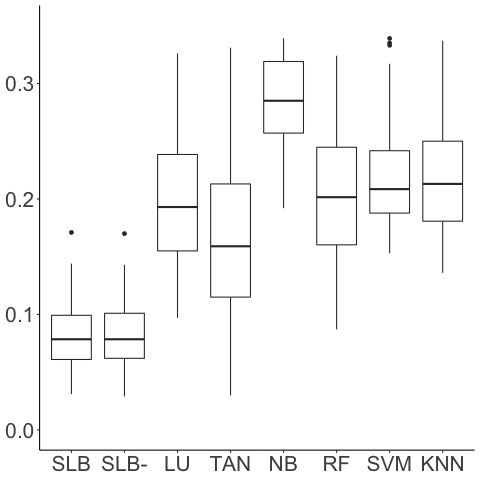} \quad &  \quad \hspace{0.1in}
    \begin{minipage}{0.5\columnwidth}
				\vspace{-3.7in}
     \caption{Misclassification error rates for the various classifiers, across 100 replicate
                  datasets. As indicated, datasets are imbbalanced  and sampled from complex BNs with common structure.}
	 \label{general_BN_complex}	
	\end{minipage}
          \end{tabular}
\end{figure}

%
%

\FloatBarrier

\begingroup
\raggedright
\small
\bibliographystyle{plainnat}
\bibliography{forest-classification}
\endgroup

\end{document}